\documentclass[lettersize,journal]{IEEEtran}
\usepackage{amsmath,amsfonts}
\usepackage{algorithmic}
\usepackage{algorithm}
\usepackage{array}
\usepackage[caption=false,font=normalsize,labelfont=sf,textfont=sf]{subfig}
\usepackage{textcomp}
\usepackage{stfloats}
\usepackage{url}
\usepackage{verbatim}
\usepackage{graphicx}
\usepackage{cite}
\usepackage{makecell}
\usepackage{multirow}
\usepackage{fancyvrb}
\usepackage{moresize}
\begin{document}

\title{A Critical Review of Common Log Data Sets Used for Evaluation of Sequence-based Anomaly Detection Techniques}

\author{Max Landauer, Florian Skopik, Markus Wurzenberger 
\IEEEcompsocitemizethanks{\IEEEcompsocthanksitem M. Landauer, F. Skopik, and M. Wurzenberger are with the Center for Digital Safety \& Security, Austrian Institute of Technology, Vienna, Austria.\protect\\
	E-mail: \{firstname\}.\{lastname\}@ait.ac.at}
}



\maketitle

\begin{abstract}
Log data store event execution patterns that correspond to underlying workflows of systems or applications. While most logs are informative, log data also include artifacts that indicate failures or incidents. Accordingly, log data are often used to evaluate anomaly detection techniques that aim to automatically disclose unexpected or otherwise relevant system behavior patterns. Recently, detection approaches leveraging deep learning have increasingly focused on anomalies that manifest as changes of sequential patterns within otherwise normal event traces. Several publicly available data sets, such as HDFS, BGL, Thunderbird, OpenStack, and Hadoop, have since become standards for evaluating these anomaly detection techniques, however, the appropriateness of these data sets has not been closely investigated in the past. In this paper we therefore analyze six publicly available log data sets with focus on the manifestations of anomalies and simple techniques for their detection. Our findings suggest that most anomalies are not directly related to sequential manifestations and that advanced detection techniques are not required to achieve high detection rates on these data sets.
\end{abstract}

\begin{IEEEkeywords}
log data analysis, anomaly detection, data sets
\end{IEEEkeywords}

\section{Introduction} \label{intro}

\IEEEPARstart{S}{ound} evaluations of machine learning algorithms are essential to validate their correct functioning, measure the accuracy of classifications, and conduct comparative studies with state of the art algorithms. Data sets are the basis for these evaluations and the selection of appropriate data sets is key to obtain representative results with general validity.

To be suitable for the purpose of evaluations, data sets are generally expected to fulfill several quality criteria \cite{kenyon2020public}, such as correctness, completeness, relevance, timeliness, realism, etc. When it comes to the evaluation of anomaly detection techniques that aim to identify rare and unexpected events in otherwise normal data \cite{chandola2009anomaly}, data sets and specifically the manifestations of anomalies must also meet the characteristics suitable for the type of detection under test. For example, anomaly detection techniques that process and analyze data instances as chronologically ordered sequences need data sets where anomalies manifest as changes in sequential patterns rather than, for example, appearance of entirely new sequence elements that could be more effectively detected by other approaches.

There is an active research community that focuses on anomaly detection in system log data. Logs create a permanent record of almost all activities that take place on a system or within an application and are therefore a valuable source of information when it comes to failure analysis or forensic investigations of cyber incidents \cite{landauer2023deep, chandola2009anomaly}. Due to their large size and repeating patterns, anomaly detection approaches are capable of capturing the normal system behavior as reflected in the log data, disclose any sudden deviations from these normal behavior models as anomalies, and trigger alerts for system operators in case that anomalous states are detected. Given that log data keeps track of the underlying workflow of applications, generated logs often form similar patterns of chronologically ordered event sequences. Logically, it is fair to assume that undesired activities (failures, attacks, etc.) also manifest as sequences and are thus detectable within the sequential patterns, such as changes of positions of certain events \cite{zhang2019robust}. Accordingly, many anomaly detection algorithms that make use of sequential patterns have been proposed in the past; specifically, deep learning methods such as Long Short-Term Memory Recurrent Neural Networks (LSTM RNNs) that are also commonly used for text processing have been widely used in recent years \cite{landauer2023deep}. Thereby, studies showed that approaches based on deep learning are generally able to outperform conventional machine learning methods such as clustering \cite{chen2021experience}.

Accordingly, it is easy to get the impression that advanced detection techniques are necessary to achieve high detection performance on log data sets that are commonly used in scientific evaluations \cite{landauer2023deep}. However, during manual analysis of these data sets, we noticed that many anomalies are either straightforward to detect or not directly related to changes of sequential patterns. Inspired by the work of Wolsing et al. \cite{wolsing2022can}, who show that SIMPLE (Sufficient, Independent, Meaningful, Portable, Local \& Efficient) detection methods achieve competitive detection rates in comparison to complex approaches using neural networks in the field of industrial control systems, we develop a set of simple yet effective and broadly applicable detection techniques and evaluate them on commonly used log data sets, including detection of previously unseen event types and sub-sequences, unusually short or long sequences, sequences with deviating event occurrence counts, changes of event ordering, and delayed event occurrences. With this experimental study we aim to answer the following two research questions: \textit{How do anomalies manifest themselves in common log data sets? What are drawbacks that render these data sets inadequate for evaluation of sequence-based anomaly detection techniques?}

We point out that this study does not recreate or compare any results from state of the art approaches, which has already been carried out in other surveys \cite{le2022log, chen2021experience, yadav2020survey}. Instead our focus lies on log data sets and their appropriateness for evaluation of anomaly detection techniques. We provide the scripts to reproduce the results presented in this paper online\footnote{Anomaly-detection-log-datasets GitHub repository available at \url{https://github.com/ait-aecid/anomaly-detection-log-datasets} (accessed 04-09-2023)}. We summarize our contributions as follows:
\begin{itemize}
	\item A review of common log data sets and their properties relevant for anomaly detection,
	\item a baseline evaluation using a set of simple detection techniques, and
	\item a critical discussion on the suitability of the data sets for evaluation of sequence-based anomaly detection approaches.
\end{itemize}
The remainder of this paper is structured as follows. Section \ref{related} provides some background of this research area and reviews related publications. Section \ref{analysis} describes the log data sets covered in this study and highlights important data properties. Section \ref{experiment} first outlines the detection techniques applied on the data sets and then provides the results of our evaluation study. Section \ref{discussion} comprises a critical discussion of our findings with respect to the appropriateness of the data sets for scientific evaluations. Section \ref{conclusion} concludes the paper.

\section{Background \& Related Work} \label{related}

\begin{figure}
	\centering
	\includegraphics[width=1\columnwidth]{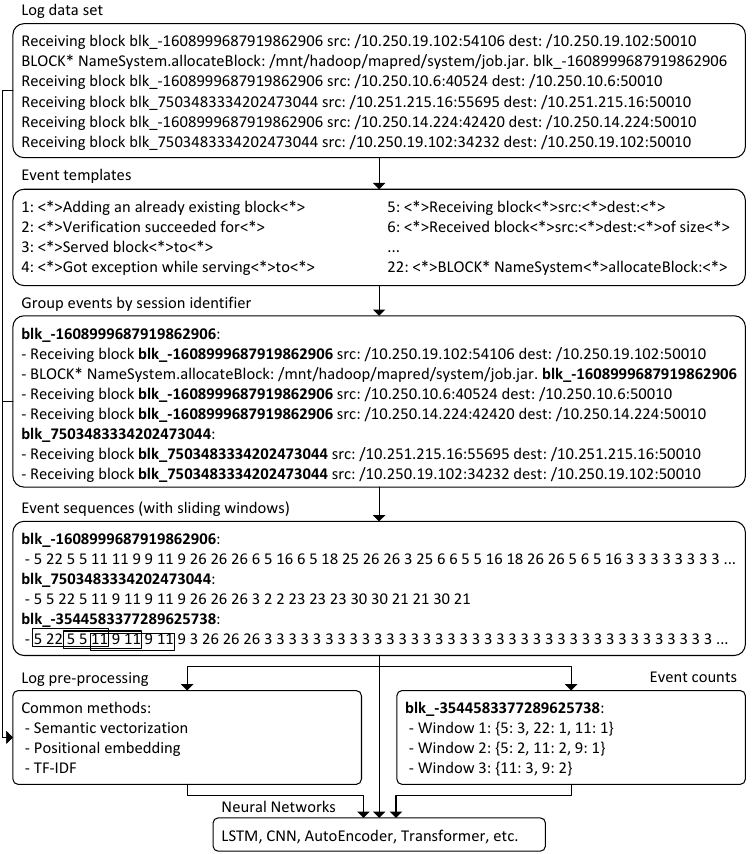}
	\caption{Workflow for anomaly detection in log data.}
	\label{fig:overview}
\end{figure}

In their survey on anomaly detection in log data with deep learning, Landauer et al. \cite{landauer2023deep} found that only a few publicly available log data sets are used by almost all of the reviewed publications. The most commonly used data sets stem from applications that produce heterogeneous log events, i.e., each log message corresponds to a specific type of event, comprising static parts and variable parameters following a specific syntax. Consider the logs in the top of Fig. \ref{fig:overview} as an example. In each log line except for the second one, ``Receiving block'' as well as ``src:'' and ``dest:'' are static, but the block identifier as well as source and destination address are variable. Event templates (sometimes also referred to as log keys or simply events \cite{chen2021experience}) are used to describe these event syntaxes and can be automatically generated by algorithms such as Drain \cite{he2017drain}. The aforementioned sample lines correspond to event type 5, while the second line in the sample logs corresponds to event type 22 that are visible in the second block in Fig. \ref{fig:overview}.

The first step in the common anomaly detection workflow depicted in Fig. \ref{fig:overview} consists of parsing the log data with templates, for the purpose of (i) assigning an event identifier to each log line and (ii) extracting parameters in a structured way. Several of the commonly used log files involve a so-called sequence identifier that can be extracted as one of the parameters and used to group events together that belong to the same process or trace. In the sample logs, this identifier corresponds to a data block (starting with ``blk\_'' and followed by a unique ID) that is processed by the application. When such identifiers are not available, sequences are sometimes also generated by sliding a window of a fixed size over the parsed data set \cite{le2022log}. Either way, the resulting sequences consist of ordered lists of event types that are represented as distinct integers in Fig. \ref{fig:overview}. Some approaches additionally apply a sliding window on the sequences (a window of size 5 with step width of 2 is displayed in the figure as an example) \cite{le2022log}, compute event counts in sequences or windows \cite{meng2019loganomaly}, transform log messages into numeric vectors with embedding techniques \cite{zhang2019robust}, apply weighting schemes such as TF-IDF \cite{yang2021semi}, use neural networks on the raw log messages \cite{hashemi2021onelog}, etc. Eventually, the data fed into anomaly detection systems usually directly relates to event sequences, as sequential patterns are assumed to be the key indicator for anomalies.

Existing surveys on sequence-based anomaly detection techniques generally focus on deep learning applications. For example, Le et al. \cite{le2022log} quantitatively compare five state of the art anomaly detection models on four data sets with focus on data pre-processing strategies, data set imbalance, and robustness to noise. Their results suggest that detection capabilities of complex models are heavily influenced by these aspects and that actually achieved detection rates are often not as good as expected as a consequence. Chen et al. \cite{chen2021experience} provide another quantitative survey that involves four unsupervised and two supervised deep learning approaches that are evaluated on two data sets. They found that anomalies incorrectly inserted in training data as well as unknown event types can strongly impact detection capabilities. In addition, they point out that conventional machine learning models such as clustering, PCA, or invariant mining, are typically more efficient than their deep learning counterparts in terms of runtime. 

Landauer et al. \cite{landauer2023deep} provide a qualitative survey of 62 state of the art approaches. Their work emphasizes the diversity of deep learning models, pre-processing strategies, and methods to transform log events into representations suitable for ingestion by neural networks. Yadav et al. \cite{yadav2020survey} provide another qualitative survey of detection approaches and discuss common challenges, model architectures, and pre-processing strategies. They also summarize the data sets used in scientific publications, but do delve into their individual properties.

The focus of aforementioned publications lies always on detection techniques; we are not aware of any works that critically analyze data sets used to evaluate these approaches. Kenyon et al. \cite{kenyon2020public} review the appropriateness of publicly available data sets in the intrusion detection domain \cite{kenyon2020public}, however, these are not the data sets typically used to evaluate sequence-based anomaly detection techniques \cite{landauer2023deep}. With this paper we therefore aim to close this gap and support researcher to better understand the data they use for evaluations.

\section{Analysis of Log Data Sets} \label{analysis}

\begin{table*}[h!]
	\setlength{\tabcolsep}{4pt}
	\scriptsize
	\caption{Overview of common log data sets}
	\label{tab:overview}
	\begin{tabular}{lccccccccccccc}
		& & \multicolumn{2}{c}{\makecell[c]{HDFS \\ (Xu et al.)}} & \multicolumn{2}{c}{\makecell[c]{BGL \\ (CFDR)}}    & \multicolumn{2}{c}{\makecell[c]{Thunderbird \\ (CFDR)}}      & \multicolumn{2}{c}{\makecell[c]{OpenStack \\ (Loghub)}}   & \multicolumn{2}{c}{\makecell[c]{Hadoop \\ (Loghub)}} & \multicolumn{2}{c}{\makecell[c]{ADFA \\ (Creech et al.)}}  \\  \cline{3-14} 
		& & Abs. & Rel. & Abs. & Rel. & Abs. & Rel. & Abs. & Rel. & Abs. & Rel. & Abs. & Rel.    \\ \hline
		Number of lines & Total & 11,197,705 & 100\% &  4,747,963 & 100\% &  211,212,192  & 100\% &  207,820 & 100\% &  394,308 & 100\% &  2,747,550 & 100\%  \\ \hline
		\multirow{3}{*}{Number of parsed events} & Total & 12,580,989 & 112.4\% &  4,747,963 & 100\% &  211,212,192  & 100\% &  57,784 & 27.8\% &  393,433 & &  2,747,550 & 100\%  \\ \cline{2-14} 
		& Normal & 12,255,230 & 97.4\% &  4,399,265 & 92.7\% &  193,744,733  & 91.7\%  &  52,290 & 90.5\% &  25,285 & 6.4\% &  2,430,162 & 88.5\%  \\ \cline{2-14} 
		& Anom. & 325,759 & 2.6\% &  348,698  & 7.3\% & 17,467,459 & 8.3\% &  5,494  & 9.5\% & 368,148  & 93.6\% & 317,388 & 11.5\% \\ \hline
		\multirow{3}{*}{Number of event types} & Total  & 33 & 100\% &  394 & 100\% &  6,178 & 100\% &  30 & 100\% &  349 & 100\% &  175 & 100\%   \\ \cline{2-14} 
		& Normal & 20 & 60.6\% &  339 & 86.0\% &  5,577 & 90.3\% &  30 & 100\%  &  229 & 65.6\% &  174 & 99.4\%  \\ \cline{2-14} 
		& Anom. & 32 & 97.0\% &  60 & 15.2\% &  652 & 10.6\% &  27 & 90.0\% &  345 & 98.9\% &  90 & 51.4\% \\ \hline
		\multirow{3}{*}{Number of sequences} & Total  &  575,061 & 100\% &  69,252 & 100\% &  5,855 & 100\% &  2,069 & 100\% &  978 & 100\% &  5,951 & 100\%  \\ \cline{2-14} 
		& Normal & 558,223 & 97.1\% &  37,823 & 54.6\% &  1,546 & 26.4\% &  1,871 & 90.4\% &  167 & 17.1\% &  5,205 & 87.5\%  \\ \cline{2-14} 
		& Anom. & 16,838  & 2.9\% & 31,429  & 45.4\% & 4,309  & 73.6\% & 198 & 9.6\% &  811  & 82.9\% & 746 & 12.5\%  \\ \hline
		\multirow{3}{*}{\makecell[l]{Number of unique \\ sequences}} & Total & 26,814  & 4.7\% & 27,571  & 39.8\% & 5,855 & 100\% &  27 & 1.3\% &  223  & 22.8\% &  3,852  & 64.7\%  \\ \cline{2-14} 
		& Normal & 21,690  & 80.9\% & 8,874 & 32.2\% &  1,546 & 26.4\% &  24 & 88.9\% &  52 & 23.3\%  &  3,122  & 81.0\% \\ \cline{2-14} 
		& Anom. & 5,133 & 19.1\% &  18,697  & 67.8\% & 4,309  & 73,6\% & 12  & 44.4\% & 195 & 87.4\% &  732  & 19.0\%  \\ \hline
		\multirow{2}{*}{\makecell[l]{Number of sequences \\ that also occur in other class}}  & Normal & 14 & 0.003\% &  0 & 0.0\% &  0 & 0.0\% &  1,842 & 98.5\% &  139 & 83.2\% &  2  & 0.04\%   \\ \cline{2-14} 
		& Anom. & 17  & 0.1\% & 0  & 0.0\% & 0 & 0.0\% &  195 & 98.5\% & 612 & 75.5\% &  6  & 0.8\%  \\ \hline
		\multirow{3}{*}{\makecell[l]{Number of unique \\ event count vectors}} & Total   & 666  & 2.5\% & 15,579  & 56.5\% & 5,855  & 100.0\% & 16 & 59.3\% &  198 & 88.8\% &  3,502  & 90.9\%  \\ \cline{2-14} 
		& Normal & 257  & 38.6\% & 4,605  & 29.6\% & 1,546  & 26.4\% & 14  & 87.5\% & 51  & 25.8\% & 2,772  & 79.2\%  \\ \cline{2-14} 
		& Anom. & 418  & 62.8\% & 10,974  & 70.4\% & 4,309 & 73.6\% &  7 & 43.8\% &  174 & 87.9\% &  732  & 20.9\% \\ \hline
		\multirow{2}{*}{\makecell[l]{Number of event count vectors \\ that also occur in other class}}  & Normal & 230  & 0.04\% & 0  & 0.0\% & 0  & 0.0\% & 1850  & 98.9\% & 143  & 85.6\% & 2  & 0.04\%   \\ \cline{2-14} 
		& Anom. & 316 & 1.9\% &  0  & 0.0\% & 0  & 0.0\% & 196  & 99\% & 619  & 76.3\% & 6  & 0.8\%  \\ \hline
		Collection date  & & \multicolumn{2}{c}{Nov. 2008} & \multicolumn{2}{c}{2005-2006} & \multicolumn{2}{c}{2005-2006} & \multicolumn{2}{c}{May 2017} & \multicolumn{2}{c}{Oct. 2015} & \multicolumn{2}{c}{$\approx$ 2013} \\ \hline
		Duration & & \multicolumn{2}{c}{$\approx$ 38.7 hours} & \multicolumn{2}{c}{$\approx$ 214 days} & \multicolumn{2}{c}{$\approx$ 244 days} & \multicolumn{2}{c}{$\approx$ 58.8 hours} & \multicolumn{2}{c}{$\approx$ 50.5 hours} & \multicolumn{2}{c}{Unknown} \\ \hline
		Distinct anomaly labels & & \multicolumn{2}{c}{1} & \multicolumn{2}{c}{43} & \multicolumn{2}{c}{33} &  \multicolumn{2}{c}{1} & \multicolumn{2}{c}{3} & \multicolumn{2}{c}{6} \\ \hline
		Label type & & \multicolumn{2}{c}{Sequences}   & \multicolumn{2}{c}{Events} & \multicolumn{2}{c}{Events}    & \multicolumn{2}{c}{Sequences}    & \multicolumn{2}{c}{Sequences}    & \multicolumn{2}{c}{Sequences}  \\ \hline
	\end{tabular}
\end{table*}

This section provides a general description of the most common log data sets used in scientific evaluations. The data sets are selected and ordered based on the survey by Landauer et al. \cite{landauer2023deep}; furthermore, we include one additional data set that we suggest as a potentially useful alternative. The following sections go over one data set after another and outline their origin and properties. Thereby, we generally refer to Table \ref{tab:overview}, which quantitatively summarizes the data sets with respect to the distribution of normal and anomalous instances.

\subsection{HDFS} \label{hdfs}

The HDFS log data set is the most frequently used data set for evaluations of anomaly detection techniques \cite{landauer2023deep} and thus the focus point of this study. The logs stem from the Hadoop Distributed File System (HDFS), which allows storage and processing of large files. Each log event contains one or more data block identifiers that enable grouping of events into sequences as discussed in the previous section. In fact, the sample logs shown in Fig. \ref{fig:overview} are taken from the HDFS data set. The main idea of detecting anomalies in this data set is that some data blocks are not processed correctly by the system, which is reflected by the log events generated as the block goes through an abnormal execution flow, in which case the entire sequence should be detected as anomalous \cite{xu2009detecting}. Note that lines containing multiple block identifiers have to be replicated for each corresponding sequence, thus the number of parsed events exceeds the number of lines (cf. Table \ref{tab:overview}).

The data set was originally collected in 2008 by Xu et al. \cite{xu2009detecting, xu2009online, xu2008mining} on a Hadoop cluster comprising more than 200 nodes. The data set was labeled by domain experts on the granularity of individual block identifiers. As stated by the original authors, labels were assigned to more than 575,000 event sequences by clustering them into event count vectors, which reduced their size to the manageable amount of only 680 unique vectors \cite{xu2009online}. Note that we obtain a total of 666 unique count vectors (cf. Table \ref{tab:overview}), likely due to the fact that publicly available event templates do not fully align with the templates used by the original authors. Even though the original logs are still available, the data set has also been provided by the log data set collection project Loghub \cite{he2020loghub}. A close inspection of that data set reveals that around 22,000 lines are missing in comparison to the original data set for unknown reasons. In addition, many authors rely on parsed versions of the HDFS data set that are available in public repositories, such as the LogDeep project\footnote{Available at \url{https://github.com/donglee-afar/logdeep}}. Given that the parsed versions lack sequence identifiers, it is difficult to ascertain their completeness. Moreover, since this version of the data set only comprises sequences of event types but lacks their timestamps, some authors incorrectly assume that timestamp information is not available even though it is present in the original logs \cite{le2022log}.

\begin{figure}
	\centering
	\includegraphics[width=1\columnwidth]{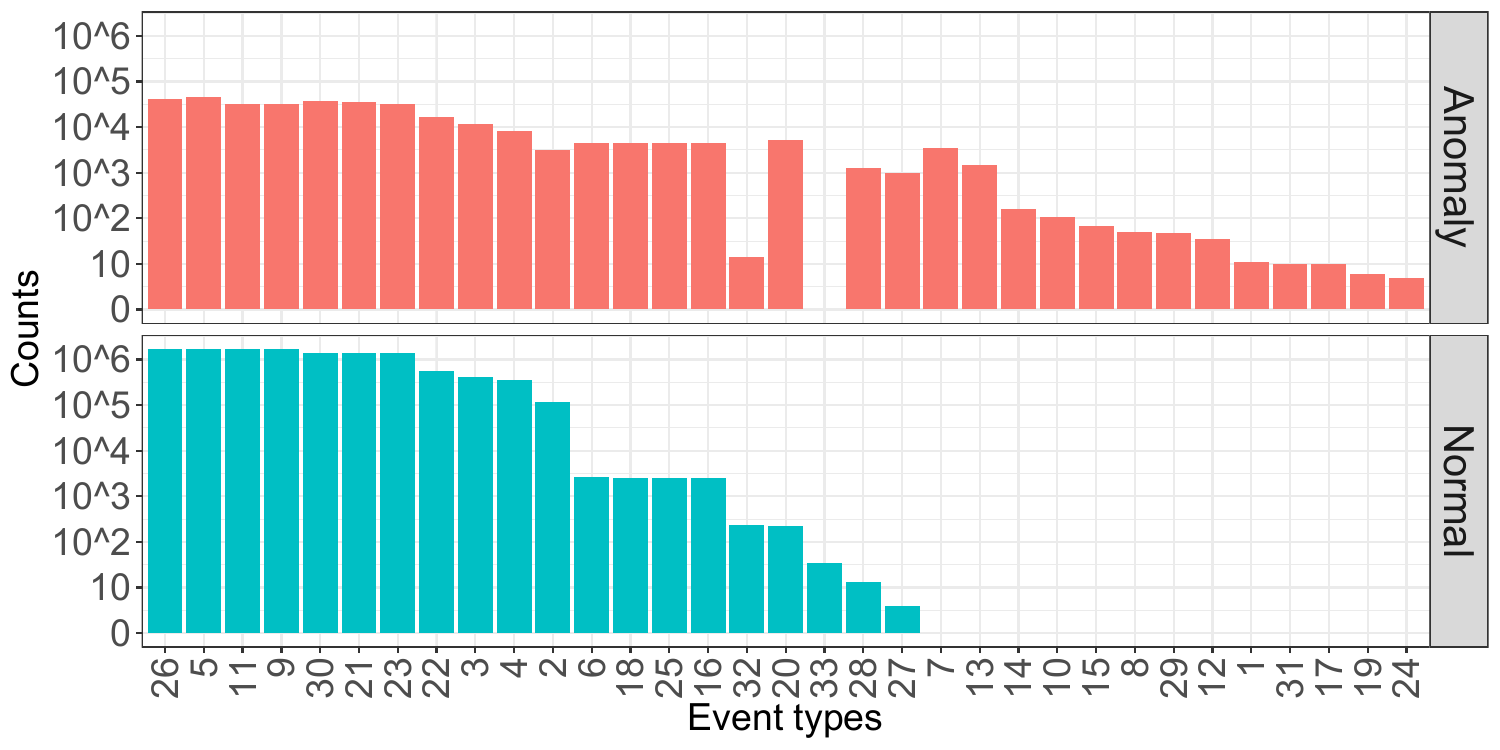}
	\caption{Event frequencies in HDFS log event sequences.}
	\label{fig:hdfs_events}
\end{figure}

Figure \ref{fig:hdfs_events} depicts the frequencies of event types represented as integer values in anomalous (top) and normal (bottom) sequences, sorted in ascending order. The plot shows that the eleven leftmost event types have a similar distribution of relative occurrence frequencies in both normal and anomalous sequences (note that their absolute numbers diverge as normal sequences are more frequent). However, many event types only occur in anomalous sequences and can thus be regarded as basic indicators for anomalies. On the other hand, event type 33 can be seen as an indicator for normal events, even though it is relatively rare, and event type 20 occurs in both normal and anomalous sequences, but is more likely to indicate anomalies as it is far more frequently part of anomalous sequences.

\begin{figure}
	\centering
	\includegraphics[width=1\columnwidth]{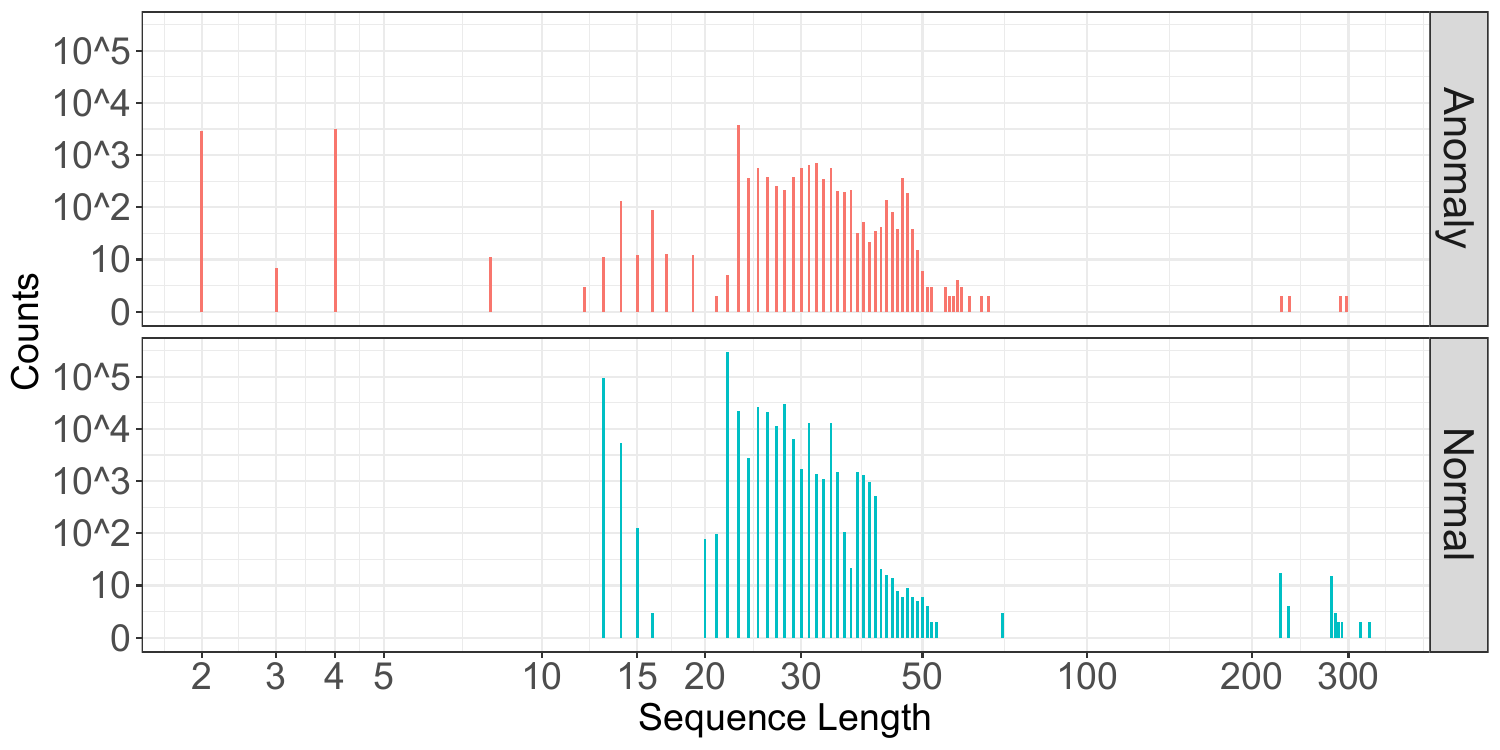}
	\caption{Distribution of HDFS log event sequence lengths.}
	\label{fig:hdfs_len}
\end{figure}

Figure \ref{fig:hdfs_len} shows that many normal and anomalous sequences also differ in length. In particular, while all normal event sequences have lengths larger than 12, more than a third of the anomalous sequences have short lengths of 2, 3, or 4.

\begin{figure}
\centering
\tiny
\begin{Verbatim}[commandchars=\\\{\},codes={\catcode`$=3\catcode`_=8}]
\textbf{Occurrence frequencies of the most common normal sequences of event types:}
\textbf{73691: 5 5 5 22 11 9 11 9 11 9 26 26 26 23 23 23 30 30 30 21 21 21}
\textbf{40156: 5 22 5 5 11 9 11 9 11 9 26 26 26 23 23 23 30 30 30 21 21 21}
\textbf{37788: 5 5 22 5 11 9 11 9 11 9 26 26 26 23 23 23 30 30 30 21 21 21}
\textbf{31213: 5 5 5 22 11 9 11 9 11 9 26 26 26}
\textbf{19311: 5 22 5 5 11 9 11 9 11 9 26 26 26}
\textbf{18297: 5 5 22 5 11 9 11 9 11 9 26 26 26}
\textbf{18078: 22 5 5 5 26 26 26 11 9 11 9 11 9 23 23 23 30 30 30 21 21 21}

\textbf{Occurrence frequencies of the most common anomalous sequences of event types:}
\textbf{1643:  5 22}
\textbf{1361:  22 5 5 7}
\textbf{1307:  22 5}
\textbf{1252:  5 5 22 7}
\textbf{612:   5 22 5 7}
\textbf{176:   22 5 5 5 26 26 26 11 9 11 9 11 9 23 23 23 30 30 30 21 21 28 26 30 21}
\textbf{158:   5 5 5 22 11 9 11 9 11 9 26 26 26 23 23 23 21 30 30 30 20 21 21}
\end{Verbatim}
\caption{Top seven most common normal (top) and anomalous (bottom) sequences and their respective occurrence counts in the HDFS data set.}
\label{fig:sample}
\end{figure}

Another important aspect to consider when evaluating anomaly detection techniques with the HDFS data set is that many sequences are identical. As stated in Table \ref{tab:overview}, the total number of 575,061 event sequences can be reduced to only 26,814 (4.7\% of total sequences) unique sequences. Even more peculiar is the fact that these unique sequences comprise only 666 (2.5\% of unique sequences or 0.1\% of total sequences) unique count vectors, i.e., lists of event frequencies that are independent from event positions in sequences. One of the reasons for this is that events that occur more or less simultaneously end up in random order. Consider the most common normal sequences displayed in Fig. \ref{fig:sample}, where the three most common sequences are identical except for the position of event type 22 among the first four events, which usually occur simultaneously according to their timestamp. Given that this effect also occurs with other event types in the sequence, the number of possible event combinations exhibited by otherwise identical sequences becomes enormous, thus resulting in the large gap between unique sequences and unique count vectors. Figure \ref{fig:sample} also shows the compositions of the most common anomalous sequences, which only comprise few elements. While event type 7 acts as an indicator for anomalies, some of these sequences only involve normal events and thus need to be detected through their short lengths. The last two sequences show that anomalous sequences generally involve the same patterns as normal sequences, but contain additional event types such as 28 or 20 that are indicators for anomalies as visible in Fig. \ref{fig:hdfs_events}. 

\subsection{BlueGene/L (BGL)} \label{bgl}

The BlueGene/L (BGL) log data set is provided by the Computer Failure Data Repository\footnote{Available at \url{https://www.usenix.org/cfdr}} (CFDR) and was originally described by Oliner et al. \cite{oliner2007supercomputers}. The log data set stems from a supercomputer located at the Lawrence Livermore National Laboratory (LLNL) that first comprises 32,768 dual-processor nodes and was upgraded to 65,536 nodes during log data collection in 2005 and 2006. We refer to the publication by Taerat et al. \cite{taerat2009blue} for a detailed technical description of the system. Similar to the HFDS log data set, Loghub \cite{he2020loghub} provides another version of the data set where few lines are missing or labeled differently for unknown reasons.

Aside from log messages, the data set contains location identifiers for components such as individual compute nodes, I/O nodes, service cards, links cards, etc., that can be used to group events into sequences. Interestingly, only some authors make use of the node identifiers to generate sequences \cite{le2022log}, while others state that it is not possible to distinguish different job executions and thus simply partition the whole data set into time windows \cite{chen2021experience}. According to our analysis, leveraging the node identifiers to group events into sequences is a reasonable approach since many similar sequences emerge, indicating that different nodes go through similar event execution flows.

\begin{figure}
	\centering
	\includegraphics[width=1\columnwidth]{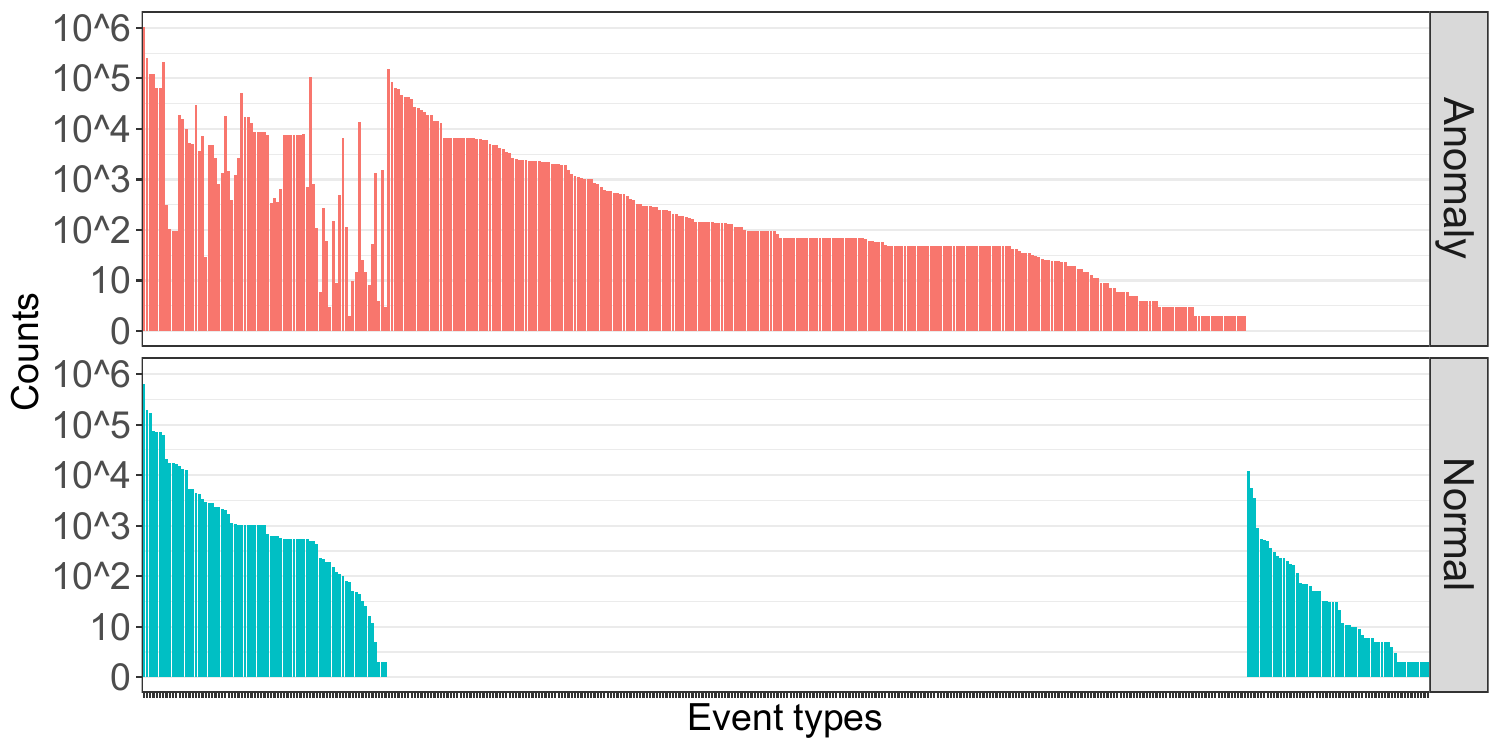}
	\caption{Event frequencies in BGL log event sequences.}
	\label{fig:bgl_events}
\end{figure}

One major difference compared to the HDFS log data set is that labels are provided on the granularity of events rather than sequences. Thereby, the labeled events form almost completely disjoint sets, i.e., the same event types are consistently labeled throughout the data set. In other words, the label of a specific event does not depend on the context of occurrence of the event but can be inferred from the event message itself. To enable comparability in Table \ref{tab:overview}, we therefore adopt the approach of existing works \cite{chen2021experience, le2022log} that tackle this issue by considering the whole sequence as anomalous if it contains at least one anomalous event. Analogous to our analysis of the HDFS file, we plot the event frequencies of normal and anomalous sequences in Fig. \ref{fig:bgl_events}, which indicates that a large fraction of events act as basic indicators for either normal or anomalous sequences.

\begin{figure}
	\centering
	\includegraphics[width=1\columnwidth]{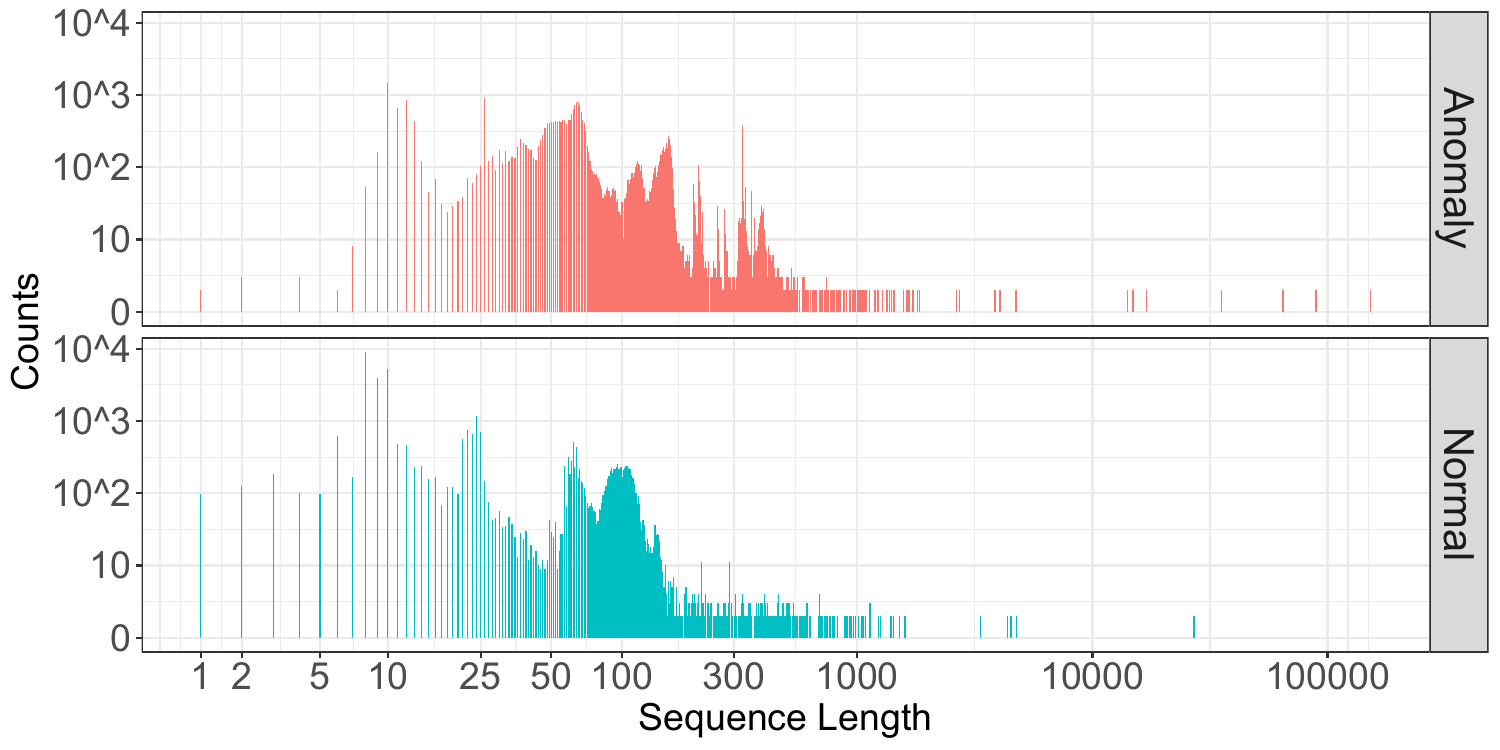}
	\caption{Distribution of BGL log event sequence lengths.}
	\label{fig:bgl_len}
\end{figure}

Figure \ref{fig:bgl_len} shows the distribution of sequence lengths in the BGL data set. Other than for the HDFS logs, there is no simple way to discern anomalous and normal sequences based on their lengths; however, the plot reveals that there are a few long anomalous sequences comprising more than 10,000 events. Given that a single labeled event in an otherwise normal sequence is enough so that the whole sequence is regarded anomalous, it stands to reason to partition long sequences into contiguous sub-sequences with individual labels. To keep the evaluation consistent across data sets, we leave the sequences unchanged for the purpose of this paper.

\subsection{Thunderbird} \label{thunderbird}

Thunderbird is yet another supercomputer log data set provided by the Computer Failure Data Repository (CFDR) and originally described by Oliner et al. \cite{oliner2007supercomputers}. The data set was collected at Sandia National Labs (SNL) at the same time as the BGL data set, but involves considerably more lines and distinct event types. Due to the high complexity of the data set, authors commonly only use a fraction of the data set, such as the first 5 million \cite{le2022log} or 20 million lines \cite{guo2021logbert}. For the purpose of this paper, we consider the whole data set for completeness. 

While there are sequence identifiers in the data, they are usually not leveraged by existing works, which resort to window- or time-based grouping \cite{le2022log}. These sequence identifiers mainly correspond to node or interface names and may be divided into groups: 77.1\% comprise two characters and digits (e.g., ``bn251''), 17.1\% comprise a period and digits (e.g., ``.741''), 4.8\% comprise complex descriptors (e.g., ``dr13iblsw2''), 0.4\% are user names (e.g., ``tbird-admin1''), 0.3\% start with ``ibr'' (e.g., ``ibr6-northern''), 0.2\% start with number signs (e.g., ``\#31\#''), and 0.1\% are IP addresses; sequence patterns have different characteristics across groups, e.g., identifiers starting with periods generally involve shorter sequences than others. As for the BGL data set, the labels of the data set are assigned to single events. Analogous to the BGL data set, our analysis shows that the events labeled as anomalous are discernible from normal events by the syntax of the log message even without considering their context of occurrence, i.e., events occurring before or after.

\subsection{OpenStack}

The OpenStack log data set was generated by the authors of DeepLog \cite{du2017deeplog}, one of the most influential papers in the research area around anomaly detection in log data \cite{landauer2023deep}. Other than the previous log data sets, the OpenStack data set was synthetically produced for the purpose of evaluating anomaly detection techniques. In particular, the authors executed a script that repeatedly carries out tasks from a pre-defined list, such as starting, stopping, pausing, and resuming virtual machines. At specific points in time while running this script, the authors injected three types of anomalies, including a timeout and errors during destruction and cleanup of virtual machines. Some of the logs contain identifiers for virtual machine instances that enable log grouping and the formation of event sequences. However, as visible in Table \ref{tab:overview}, only 27.8\% of all lines contain such an identifier; all other lines are omitted from our analysis. Unfortunately, the original data set seems to be no more available. We therefore resort to the version provided by Loghub \cite{he2020loghub}.

\begin{figure}
	\centering
	\includegraphics[width=1\columnwidth]{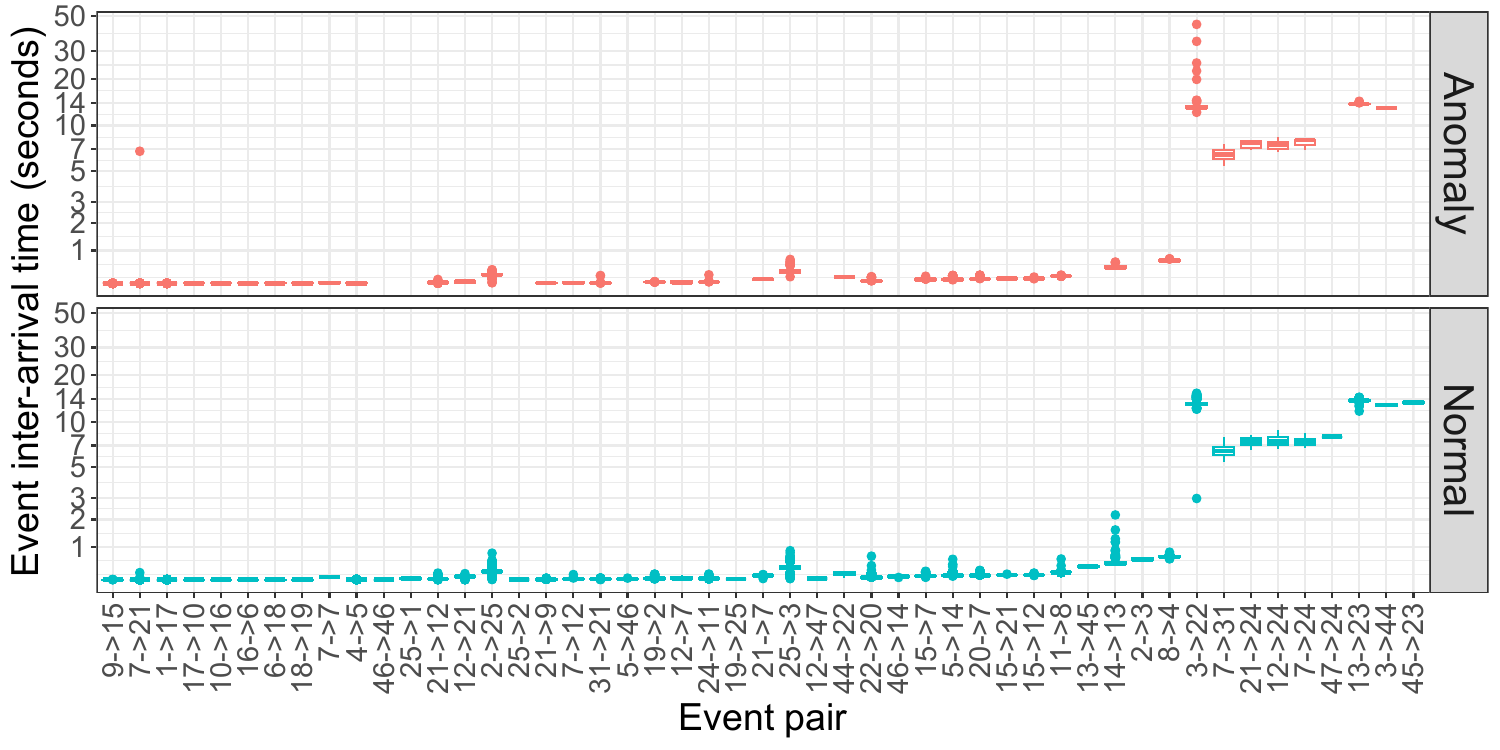}
	\caption{Event inter-arrival times in the OpenStack data set.}
	\label{fig:interarr}
\end{figure}

As Table \ref{tab:overview} shows, there is a high overlap of 98.5\% between normal and anomalous sequences, i.e., these sequences are identical. This renders application of anomaly detection techniques that only consider event sequences without contextual information infeasible. This fact by itself is not surprising: According to the original authors, anomalies should manifest by the time taken to build instances, which is reflected by the inter-arrival time of events in sequences; however, this does not seem to be the case for the log data at hand. Figure \ref{fig:interarr} depicts the distribution of inter-arrival times between all consecutive events as boxplots for normal and anomalous sequences, only to reveal that there are no significant deviations with respect to event inter-arrival times, specifically the times to reach events corresponding to starting a virtual machine (event type 22), stopping a virtual machine (event type 23), and deleting a virtual machine (event type 24). Only 5 out of 198 anomalous sequences show an increase of inter-arrival time between event type 3 and event type 22. Similar issues regarding anomalies have also been observed by other authors using the OpenStack data set \cite{kalaki2023anomaly, hashemi2021onelog}. Therefore, we do not further investigate the data set in this study.

\subsection{Hadoop} \label{hadoop}


Similar to the OpenStack data set, the Hadoop data set was synthetically generated in a lab environment to evaluate log sequence clustering \cite{lin2016log}. The authors follow the approach from Shang et al. \cite{shang2013assisting} and execute WordCount and PageRank applications on a Hadoop cluster. After some normal runs, the authors manually inject three distinctly labeled anomaly cases by turning off the server (machine down), disconnecting the server (network disconnected), and filling up the hard disk (disk full). While the original data set is not available anymore, a Hadoop data set containing the same anomalies is provided by Loghub \cite{he2020loghub}. Other than the previous data sets, logs from each run of an application are placed in separate files, which makes it easy to group events into sequences.

\begin{figure}
	\centering
	\includegraphics[width=1\columnwidth]{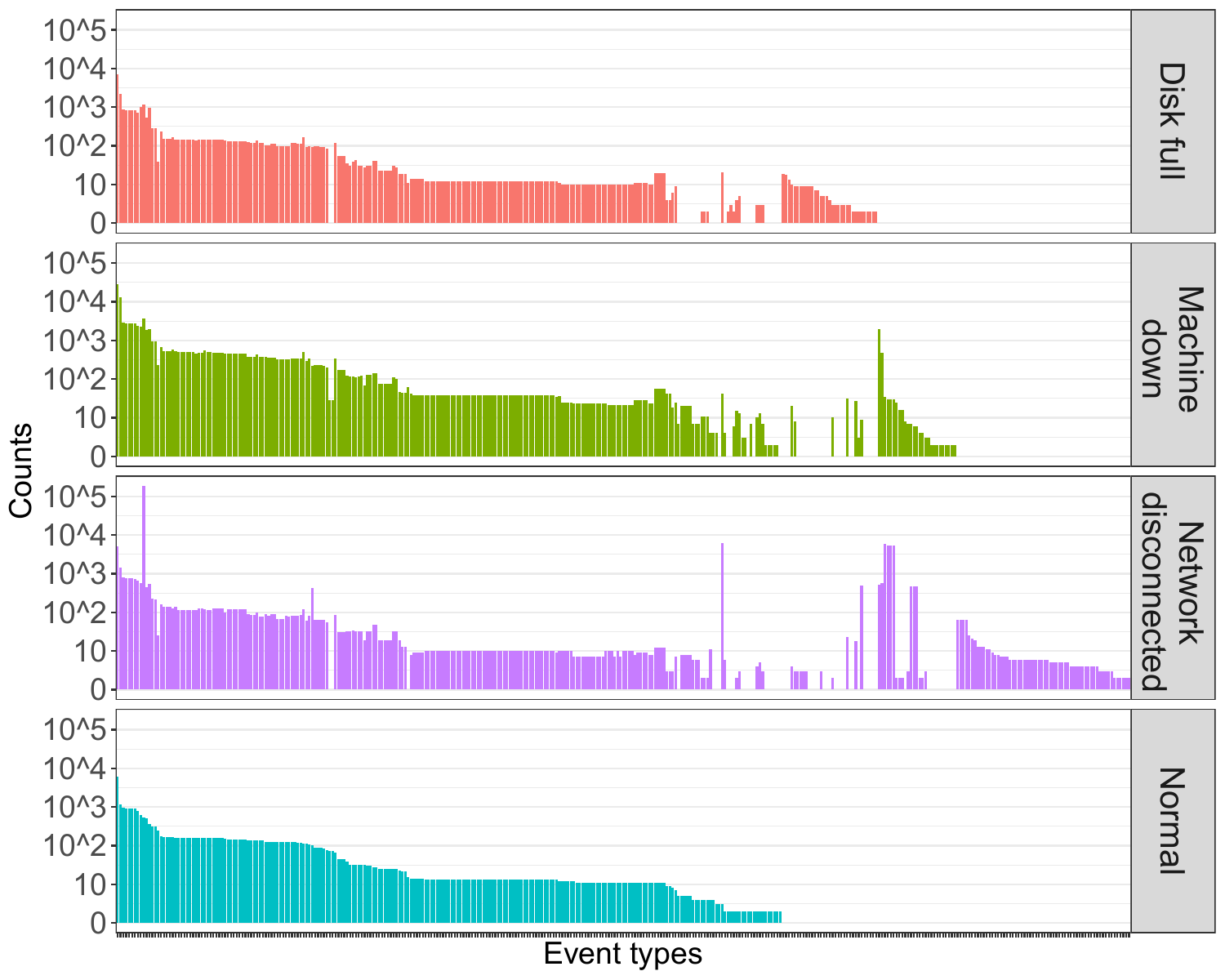}
	\caption{Event frequencies in Hadoop log event sequences.}
	\label{fig:hadoop_events}
\end{figure}

Another similarity to the OpenStack data set is that there is a high overlap between normal and anomalous sequences, with 83.2\% of normal sequences having at least one identical counterpart in the anomalous set and 75.5\% of anomalous sequences being identical to one or more normal sequences. Accordingly, analysis of sequences alone is not an adequate approach for anomaly detection. Figure \ref{fig:hadoop_events} shows that the overall distribution of event frequencies is similar across all classes, except for few anomalous sequences that involve new event types that do not occur in the normal case. Closer inspection shows that many of the events only occurring in anomalous classes are the result of a single application run, for example, the log message ``Failed to renew lease'' is printed every single second in affected applications. Given that most sequences are identical, it is also not possible to recognize anomalies by their sequence lengths. We manually compared the event parameters of common identical sequences but were also unable to identify any differences that discern normal from anomalous instances. Despite these problems, we use this data set in our evaluation study to emphasize issues with misleading evaluation metrics.

\subsection{ADFA}

The Australian Defence Force Academy Linux Dataset (ADFA-LD) was generated by Creech et al. \cite{creech2013generation} in 2013 to overcome issues with log data sets that were commonly used for evaluation of intrusion detection techniques at that time. Other than the aforementioned data sets that focus on system failures, the ADFA data set makes use of cyber attacks to generate anomalies in log data. In particular, the authors of the data set created a test environment with web applications vulnerable to known attacks and collected low-level system call logs during normal operation (e.g., web browsing or document processing) as well as execution of six attack cases.

\begin{figure}
	\centering
	\includegraphics[width=1\columnwidth]{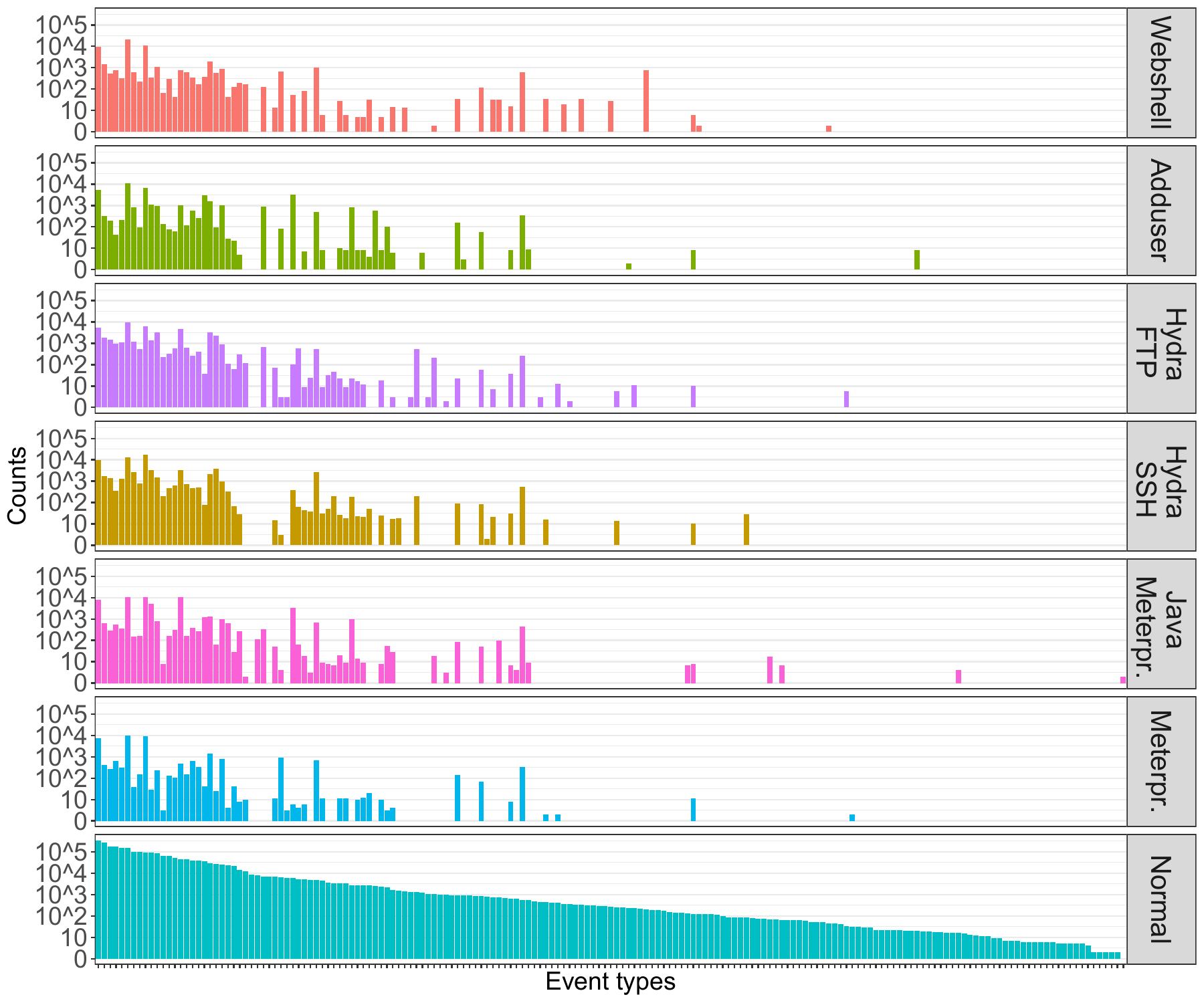}
	\caption{Event frequencies in ADFA log event sequences.}
	\label{fig:adfa_events}
\end{figure}

Figure \ref{fig:adfa_events} shows that all event types that occur in the data set also occur in sequences of the normal class, i.e., there are no event types that act as indicators for specific attacks, except for a single occurrence of event type 173 occurring during the ``Java Meterpreter'' attack. The overall event frequency distributions show that the system calls generated by attacks differ from the event frequencies during normal operation, which suggests that detection of anomalies and even classification of attack cases is feasible. We also analyzed the sequence lengths and confirm that anomalous sequences are not shorter or longer than normal ones.

Even though this data set is rarely used for evaluation of anomaly detection techniques (not a single publication considered this data set in the survey by Landauer et al. \cite{landauer2023deep}), we add the data set to our evaluation study in an attempt to propose a new data set for future evaluation that overcomes issues with commonly used data sets. In particular, using system calls avoids the need for parsing since the operations are represented as distinct integer numbers and thus reduces the influence of parsing on the detection accuracy. Moreover, system calls are generally ordered in a consistent way and are therefore not affected by permutations of simultaneously occurring events as it is the case in the HDFS log data set.

\subsection{Data Set Complexity}

We compare the data sets described in the previous sections with respect to the complexities of their sequential patterns. To this end, we apply measures for entropy and repetitiveness.

\subsubsection{Entropy}

\begin{figure}
	\centering
	\includegraphics[width=1\columnwidth]{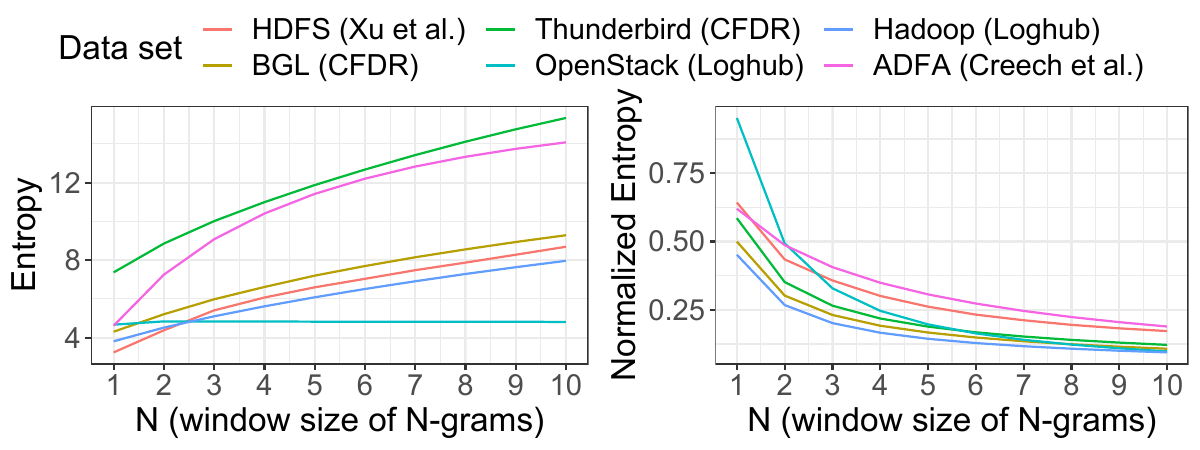}
	\caption{Entropy (left) and normalized entropy (right) of N-grams.}
	\label{fig:entropy}
\end{figure}

We use entropy to measure how evenly distributed certain parts of the sequences (i.e., contiguous sub-sequences) are across the data set, where low entropy indicates that some of these parts are occurring much more often than others and high entropy corresponds to a more random distribution. Since entropy does not account for sequential ordering, we leverage $N$-grams of various sizes and compute the entropy for each $N$ separately. To compensate for the varying numbers of distinct event types in the data sets, we also compute the normalized entropy that bases on the maximum entropy that is reached if all $N$-grams occur the same number of times. Figure \ref{fig:entropy} shows that the Thunderbird and ADFA data sets yield the highest total entropy, while the entropy computed for OpenStack data set is comparatively low and does not increase for higher $N$. Considering also the normalized entropy, we see that the high total entropy of the Thunderbird data set is caused by the large number of distinct events. The ADFA data set yields the highest normalized entropy and is closely followed by the HDFS data set, especially for large $N$. The normalized entropy of the OpenStack data set on the other hand starts out with a comparatively high value for $N=1$, meaning that single event frequencies are more evenly distributed than in other data sets, but quickly diminishes for larger $N$.

\subsubsection{Repetitiveness}

\begin{figure}
	\centering
	\includegraphics[width=1\columnwidth]{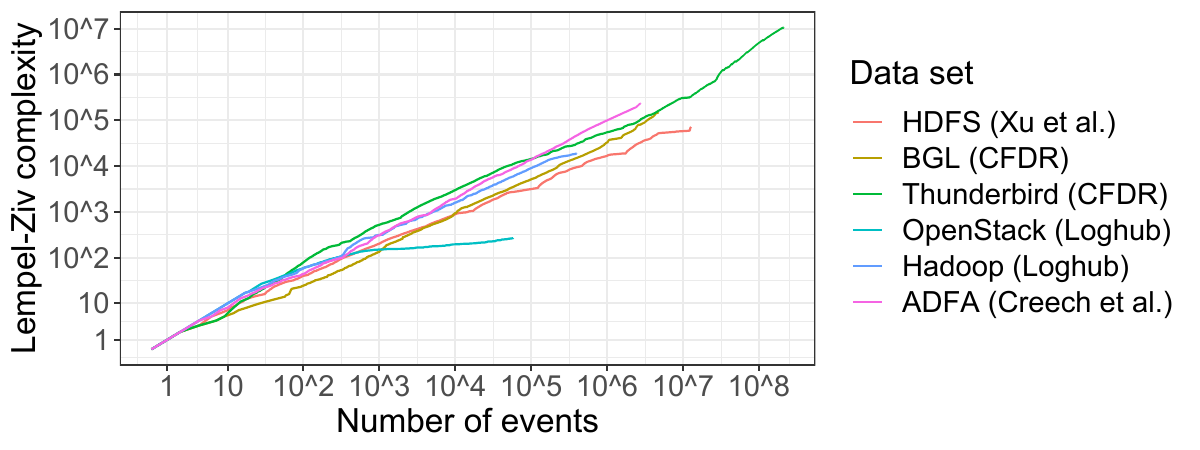}
	\caption{Sequence repetitiveness measured with Lempel-Ziv complexity.}
	\label{fig:lz}
\end{figure}

Arguably, the number of distinct event types appearing in a data set is an indicator for its complexity as larger numbers of events have the potential to form more diverse patterns (the numbers of distinct events are stated in Table \ref{tab:overview}). To assess whether they actually form such complex sequences or instead occur in the same patterns that repeat over and over, we leverage the Lempel-Ziv complexity \cite{lempel1976complexity}. In short, this measure counts the number of different contiguous sub-sequences encountered when processing all sequences from beginning to end, where already observed sub-sequences are stored in a dictionary for all sequences in a data set. Figure \ref{fig:lz} plots the Lempel-Ziv complexity with respect to the total number of events within each consecutively processed sequence. The plot shows that all data sets except for the OpenStack data set roughly exhibit the same level of complexity, with the HDFS data set ending up with a slightly lower complexity than the other data sets. The OpenStack data set, however, breaks out of the overall trend at around 1,000 processed events, after which most of the subsequently analyzed sequences contain many already observed sub-sequences, causing that the total complexity levels off.

\section{Evaluation Study} \label{experiment}

In this section we evaluate how anomalies manifest in common log data sets. We first outline the setup of our experiment and then briefly describe every detection technique we apply on the data sets before summarizing the results. 

\subsection{Setup} \label{setup}

The purpose of our study is to assess the appropriateness of five of the previously described data sets (HDFS, BGL, Thunderbird, Hadoop, and ADFA) for evaluating anomaly detection techniques leveraging event sequences. We therefore design an experiment that evaluates simple detection mechanisms on the data sets and measures whether they are sufficient to achieve competitive detection rates. Comparing different detection mechanisms with each other then allows us to better understand how anomalies manifest themselves in each data set. We ensure that the selected detection techniques are as generic as possible to be applicable to all sorts of data sets and simple enough so that it is easy to understand why specific instances are reported as anomalies. Similar to common deep learning methods \cite{landauer2023deep}, our detection techniques also only process grouped event type sequences without any contextual information other than the timestamp. 

The focus of our evaluation study lies on semi-supervised detection, where only instances of a fraction of the normal class are available for training. This is the most common scenario for anomaly detection as anomalies generally correspond to unexpected or unusual system behavior that is non-trivial to define in advance \cite{landauer2023deep}. We follow the strategy from Du et al. \cite{du2017deeplog} and randomly sample 1\% of the normal sequences for training, except for the Hadoop data set where we use 10\% since only 167 normal sequences are available. After training, we run the detector on the test data, which comprises the remaining normal sequences as well as the anomalous sequences, and measure its ability to discern these two classes. In particular, we count true positives (TP) as correctly detected anomalous sequences, true negatives (TN) as correctly undetected normal sequences, false positives (FP) as incorrectly detected normal sequences, and false negatives (FN) as incorrectly undetected anomalous sequences. Based on these counts we then compute precision ($Prec = \frac{TP}{TP + FP}$), recall or true positive rate ($Rec = TPR = \frac{TP}{TP + FN}$), true negative rate ($TNR = \frac{TN}{TN + FP}$), and F1 score ($F1 = \frac{2 \cdot Prec \cdot Rec}{Prec + Rec}$). We repeat sampling and evaluation 25 times for each data set and each detector to also capture the variance of our results. Our detection techniques rely either on none or on a single threshold for detection. In the following, we iterate over all thresholds between 0 and 1 in steps of 0.01 and select the one that maximizes F1. We investigate the influence of the threshold separately in Sect. \ref{thresh}. Given that two of our data sets - BGL and Thunderbird - are labeled on the granularity of events rather than sequences, we additionally evaluate them by computing aforementioned metrics accordingly and present the results in Sect. \ref{eventbased}.

\subsection{Detection Techniques}

This section describes each detection technique that is applied on the selected data sets.

\subsubsection{New Event Types}

The detection for new event types monitors all events that appear in any sequence of the training data set to build a model of distinct event types that are known to occur during normal operation. In the detection phase, all sequences that contain one or more event types that are not known from the training phase are reported as anomalies. This is the most basic detection technique but has been effectively used in intrusion detection systems such as the AMiner \cite{landauer2023aminer}. It is expected to work well in data sets where normal and anomalous event types resemble disjoint sets, such as the BGL and Thunderbird data set, but also achieve good performance in data sets with many event types that act as indicators for anomalies, such as the HDFS data set.

\subsubsection{Deviating Sequence Lengths}

This detection technique learns the minimum and maximum sequence lengths of all sequences in the training data. Subsequently, all sequences in the test data set with lengths shorter than the minimum or longer than the maximum are reported as anomalies. This detection technique specifically aims to recognize unusually short sequences in the HDFS data set (cf. Fig. \ref{fig:hdfs_len}). We also consider the combination of this technique with the detection for new events, so that sequences that are reported by either one of those two techniques are considered anomalies.

\subsubsection{Event Count Vector Clustering (ECVC)}

The idea behind this detection technique is that normal sequences that occur in the test data are similar to one or more sequences in the training data in terms of event types and their respective frequencies, while anomalous sequences are dissimilar to every sequence of the training data. For this purpose, we create event count vectors for sequences, where each vector index corresponds to a specific event type and the value at that index reflects the number of times the event type occurs in a sequence. After transforming all training sequences into count vectors, we use the $L_1$ norm\footnote{Distance between vectors $a$ and $b$ is computed by $\sum_i \left| a_i - b_i \right| $.} as a similarity metric that classifies count vectors from the test data set as anomalous if their similarity to all of the training instances is lower than a threshold. Note that we use the $L_1$ norm due to the fact that it handles high-dimensional data better than higher-order norms \cite{aggarwal2001surprising} and has already been applied with count vectors for anomaly detection in user behavior patterns \cite{landauer2022user}. In the following, we also consider a variant of this method where the count vector indices are weighted higher when the corresponding event types only appear in few sequences, analogous to the well-known TF-IDF measure \cite{schutze2008introduction}. In addition, we combine this technique with detection based on new events and sequence lengths in the following, where sequences are detected as anomalous if any of the combined techniques reports them as such.

\subsubsection{N-grams}

This detection technique runs a sliding window of size $N$ at a step width of 1 over all training sequences and learns the ordered sub-sequences of event types inside the window (note that the term sub-sequences always refers to contiguous chunks of sequences in this paper). In the detection phase, a window of the same size slides over the test sequences and the sub-sequences inside the window are compared with the ones known from training. Forrest et al. \cite{forrest1996sense} proposed to count the number of mismatching sub-sequences and determine the whole sequence as anomalous if the normalized count exceeds a certain threshold. The optimal value for $N$ is non-trivial to determine and is highly dependent on the data. In our experiments, we therefore use 2, 3, and 10 as values for $N$ as these window sizes have shown to be useful choices in other works \cite{creech2013generation, du2017deeplog, landauer2023deep, forrest1996sense}. 

\subsubsection{Edit distance}

This detection technique is similar to the ECVC in the sense that it computes a distance between a test sequence and every training sequence and classifies it as anomalous if no sufficiently similar pair is found. However, other than the ECVC that makes use of unordered count vectors, this detector computes the normalized edit distance, which counts the number of insertions, deletions, and replacements of event types to transform one sequence into another, which inherently relies on the ordering of event types in sequences \cite{schutze2008introduction}. As such, this detection technique is able to detect anomalies that manifest as additional or missing event types as well as changes of event ordering. Accordingly, this detection technique should work better than ECVC for anomalies with sequential manifestations, since ECVC only considers event occurrences independent from their position in the sequences.

\subsubsection{Event Timing}

Most activities that occur in specific states of processes take a certain amount of time that remains relatively steady or at least within a certain range throughout multiple executions. This is reflected in the inter-arrival times of log events that usually mark the start, stop, or intermediate steps of such processes. For example, consider the inter-arrival times displayed in Fig. \ref{fig:interarr}, where the time to start a virtual machine is always in the range of 12 to 15 seconds. The assumption of this detection method is that in some anomalous executions, the event types remain the same, but the activities carried out in between take much longer or shorter. For our simple detection method we therefore compute the minimum and maximum passed time between each pair of consecutive event types in the training data set and then classify test sequences as anomalies if the inter-arrival time between any of the involved event pairs deviates too much from the range learned specifically for this event pair, i.e., the relative difference to the range boundary exceeds a certain threshold. We opted against statistical tests (e.g., for normal distributions) on the inter-arrival times since some event pairs occur few or even just a single time in the training data.

\subsection{Results}

This section summarizes the results obtained from sequence- and event-based detection on the selected data sets.

\subsubsection{Sequence-based detection} \label{seq}

\begin{table*}[]
	\scriptsize
	\centering
	\caption{Average and maximum (in brackets) of F1 scores}
	\label{tab:results}
	\begin{tabular}{lcccccc}
		& \makecell[c]{HDFS \\ (LogDeep)} & \makecell[c]{HDFS \\ (Xu et al.)} & \makecell[c]{BGL \\ (CFDR)} & \makecell[c]{Thunderbird \\ (CFDR)} & \makecell[c]{Hadoop \\ (Loghub)} & \makecell[c]{ADFA \\ (Creech et al.)} \\ \hline
		Event                       & 63.3 & 53.9 (79.0) & \textbf{98.8} (99.4) & 92.4 (94.3) & 28.8 (32.7) & 22.1 (33.2) \\ \hline
		Length                      & 54.1 & 56.0 (59.0) & 2.8 (12.1) & 65.1 (96.6) & 19.4 (50.8) & 2.9 (6.3) \\ \hline 
		Event + Length              & 90.4 & 72.0 (92.7) & 98.7 (99.4) & 91.9 (94.0) & 33.5 (53.7) & 22.4 (32.8) \\ \hline 
		ECVC                        & 96.0 & 96.0 (96.6) & 93.5 (94.0) & 94.7 (97.0) & \textbf{91.5} (91.5) & 33.9 (46.5) \\ \hline 
		\makecell[l]{Event + Length + ECVC}       & \textbf{98.8} & 96.4 (98.7) & 98.7 (99.4) & 91.9 (94.0) & \textbf{91.5} (91.5) & 31.6 (37.9) \\ \hline
		ECVC (idf)                  & 96.3 & 96.5 (97.5) & 93.7 (94.3) & 94.7 (97.0) & \textbf{91.5} (91.5) & 31.0 (38.9) \\ \hline 
		\makecell[l]{Event + Length + ECVC (idf)} & 96.9 & \textbf{97.1} (98.4) & 98.7 (99.4) & 91.9 (94.0) & \textbf{91.5} (91.5) & 30.6 (38.2) \\ \hline
		2-gram                      & 85.2 & 86.2 (95.9) & 92.3 (93.0) & 84.9 (84.9) & \textbf{91.5} (91.5) & 27.9 (30.3) \\ \hline 
		2-gram + Length             & 95.1 & 87.8 (95.6) & 96.1 (96.9) & 84.9 (84.9) & \textbf{91.5} (91.5) & 26.5 (29.0) \\ \hline
		3-gram                      & 81.5 & 85.3 (91.0) & 62.7 (62.7) & 84.9 (84.9) & \textbf{91.5} (91.5) & 28.4 (29.6) \\ \hline 
		10-gram                     & 33.5 & 5.7 (5.7) & 62.7 (62.7) & 84.9 (84.9) & \textbf{91.5} (91.5) & 28.2 (28.8) \\ \hline 
		Edit                        & 71.8 & 71.6 (72.0) & 93.4 (93.6) & \textbf{95.2} (97.4) & \textbf{91.5} (91.5) & \textbf{34.3} (44.0) \\ \hline 
		\makecell[l]{Event + Length + Edit}       & 97.8 & 83.3 (98.0) & 98.7 (99.7) & 91.9 (94.0) & \textbf{91.5} (91.5) & 32.4 (36.2) \\ \hline
		Event Timing                & - & 45.0 (53.2) & 81.6 (82.5) & 89.1 (95.5) & 89.1 (91.4) & - \\ \hline \hline
		Le et al. \cite{le2022log} & 96.6 & - & 99.9 & 76.0 & - & - \\ \hline
		Guo et al. \cite{guo2021logbert} & 82.3 & - & 90.4 & 96.64 & - & - \\ \hline
		Yang et al. \cite{yang2021semi} & 98.8 & - & 98.2 & - & - & - \\ \hline
		Chen et al. \cite{chen2021experience} & - & 94.5 & 96.7 & - & - & - \\ \hline
		Meng et al. \cite{meng2019loganomaly} & - & 95.0 & 96.0 & - & - & - \\ \hline
		Catillo et al. \cite{catillo2022autolog} & - & - & 95.0 & - & 97.0 & - \\ \hline
		\makecell[l]{Sudqi Khater et al. \cite{sudqi2019lightweight}} & - & - & - & - & - & 94.9 \\ \hline
	\end{tabular}
\end{table*}

\begin{figure*}[ht]
	\centering
	\includegraphics[width=1\textwidth]{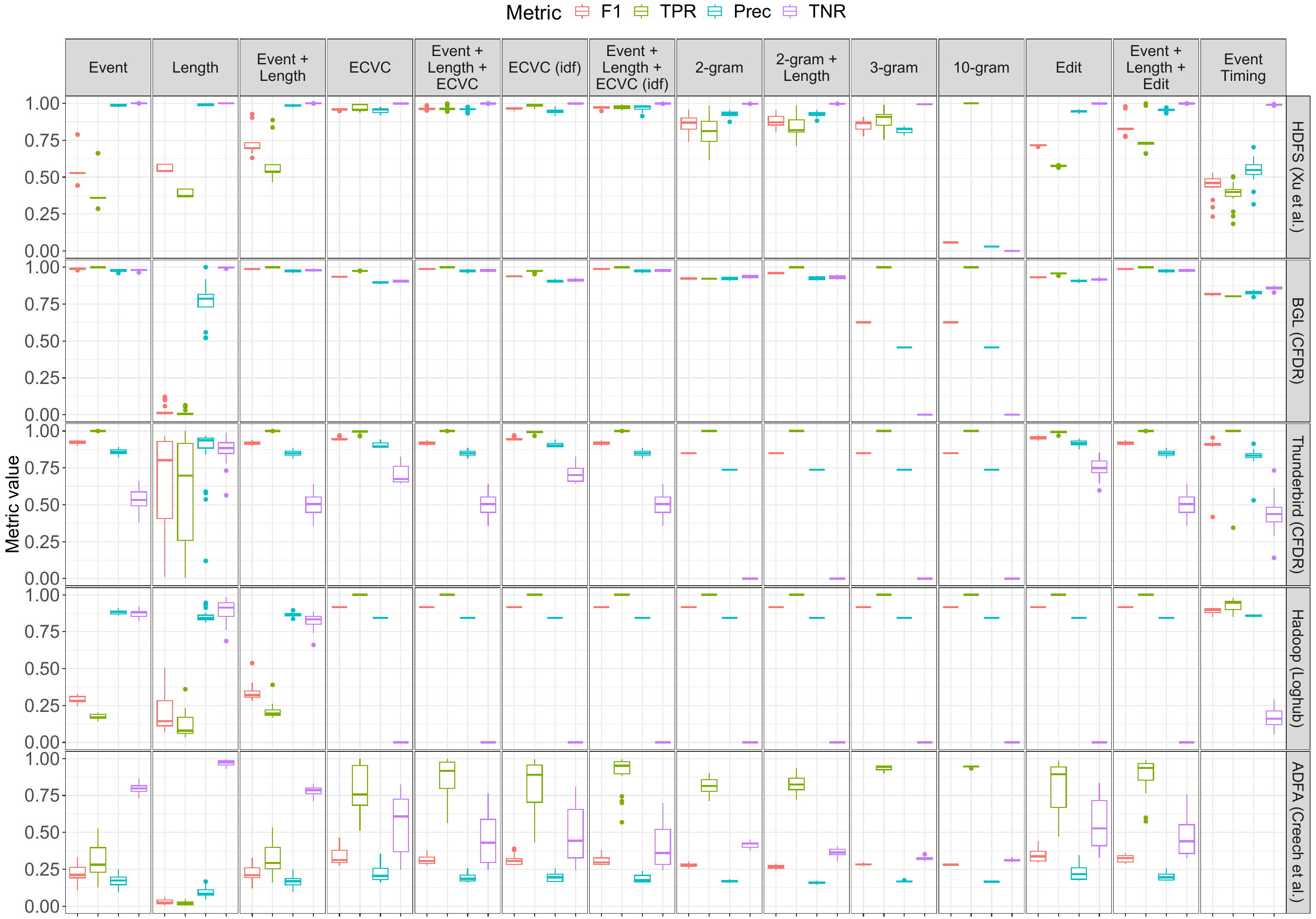}
	\caption{Evaluation results for sequence-based detection techniques on HDFS, BGL, Hadoop, and ADFA data sets.}
	\label{fig:fone}
\end{figure*}

We present the evaluation results of 25 runs with randomly sampled training sequences in Table \ref{tab:results}, which contains the average and maximum (in brackets) F1 scores obtained by applying aforementioned detection techniques and their combinations on the data sets. Highest average scores achieved for each data set are in bold. In the bottom of the table, we also provide some benchmark results that have been reported in state of the art surveys and publications on semi-supervised anomaly detection. We point out that comparability is limited as these works use different splits between training and test data sets and often only report the best scores achieved from multiple runs \cite{chen2021experience}. Note that for the HDFS data set, we use both the pre-processed version from the LogDeep repository as well as the original data set provided by Xu et al. \cite{xu2008mining} since the obtained results do not coincide. In particular, the combination of detection based on new events and sequence lengths yields an F1 score of 90.4\% on the LogDeep version, but only an average of 72.0\% on the original data set. A closer inspection of the reason for this peculiarity is that the training sequences in the LogDeep version of the HDFS data set do not involve event type 20, which is an indicator for anomalies (cf. Sect. \ref{hdfs}) but also occurs in 219 (0.04\%) of the normal sequences and thus deteriorates detection performance if randomly drawn into the set of training sequences. This effect shows the importance of repeating evaluations with multiple randomly drawn samples to obtain representative and comparable results.

Figure \ref{fig:fone} provides a boxplot for a more detailed view on the results that also includes TPR (identical to recall), precision, and TNR in addition to the F1 score. The plot confirms that for the HDFS data set, more than half of all anomalies are very simple to detect based on the combination of new event types and sequence lengths. ECVC detection further improves the scores and achieves competitive performance compared to advanced deep learning approaches (cf. Table \ref{tab:results}). Another interesting observation is that ECVC generally yields better scores compared to edit distance and N-gram based detection, which indicates that event position in sequences is less relevant for detection than event occurrence frequencies, and that random event orders due to simultaneous occurrence has an adverse effect on the detection rates. IDF weighting appears to have a positive influence on the ECVC detection, which is reasonable as relevant event types such as event type 20 receive a higher weight compared to event types that occur in many sequences and are not related to anomalies. 

The results obtained for the BGL data set show that simple detection of new event type occurrences is sufficient to obtain competitive results. This is intuitively reasonable, since the normal and anomalous event types are almost completely disjoint sets as stated in Sect. \ref{bgl}. A sample size of 1\% appears to be sufficient to train the detector for almost all normal event types and avoid many false alarms. Interestingly, the best results for the Thunderbird data set are achieved by edit distance detection, even though ECVC and event based detection only fall short by a small margin. This indicates that event order is a relevant factor when discerning normal from anomalous sequences in this data set.

Regarding the Hadoop data set, the results for F1 (91.5\%), TPR (100\%), and precision (84.4\%) indicate adequate detection performance at first glance. However, TNR of 0\% reveals that in fact the detector just reports every single sequence as anomalous and thus the detection results are of no practical value. Clearly, no detection technique is able to yield good results due to the fact that many normal and anomalous sequences are identical as described in Sect. \ref{hadoop}. This demonstrates the importance of considering metrics such as TNR in addition to F1, precision, and recall, to avoid misleading results in imbalanced data sets such as the Hadoop data set where 82,9\% of all sequences are anomalous \cite{le2022log}.

The ADFA data set appears to be more challenging in comparison to other data sets. Specifically, the achieved precision is relatively low across all detection techniques and there is no significant difference between techniques that leverage sequence ordering and those that focus on event frequencies. Note that log events do not include timestamps and thus detection based on event timing is omitted from the plot. We suggest this data set as a useful candidate for future evaluations of sequence-based anomaly detection techniques that improve upon our baseline results.

\subsubsection{Influence of detection thresholds} \label{thresh}

\begin{figure}
	\centering
	\includegraphics[width=1\columnwidth]{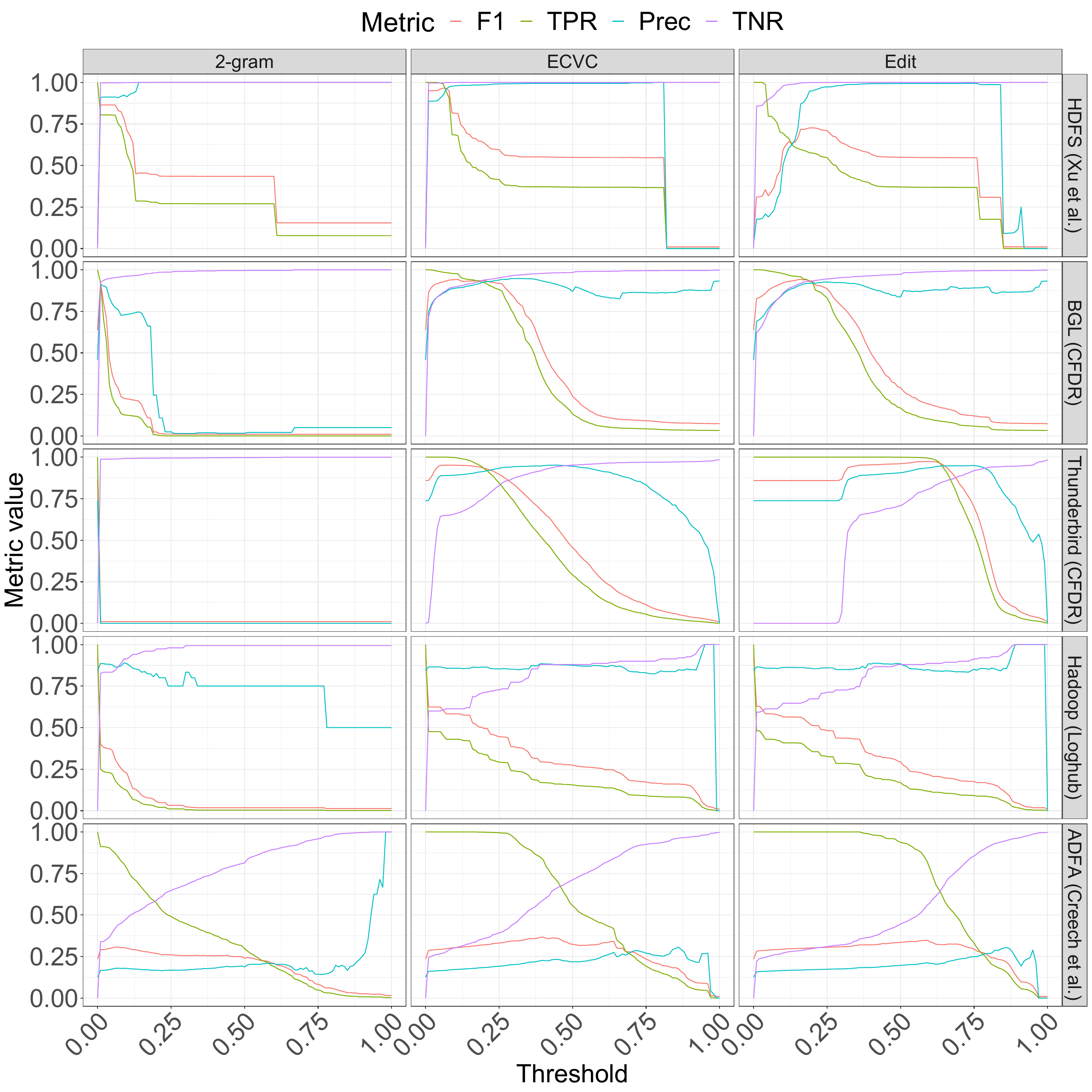}
	\caption{Influence of the detection threshold on evaluation metrics.}
	\label{fig:thresh}
\end{figure}

As stated in Sect. \ref{setup}, we optimized the results in the previous section by fine-tuning the detection thresholds to maximize F1. In practice, however, such a fine-tuning is not feasible due to the lack of a ground truth. We therefore investigate the parameter influence on the results of 2-gram, ECVC, and edit distance detection, by iterating the threshold in the range 0 to 1 in steps of 0.01. Figure \ref{fig:thresh} shows the progression of evaluation metrics for each combination of data set and detection technique from a single evaluation run. For the HDFS data set, ECVC is clearly the most preferable detection technique and yields high F1 scores for thresholds lower than 0.1. Both ECVC and 2-gram achieve nearly perfect TNR for any thresholds, indicating that almost all normal sequences are straightforward to classify as such. Regarding the BGL data set, both ECVC and edit distance perform well for thresholds smaller than 0.25 and yield high precision independent from the threshold, while 2-gram detection shows a much narrower band where adequate results are achieved. On the Thunderbird data set, no true positives are detected by the 2-gram method for any threshold larger than 0, because a few very dissimilar sequences cause that almost all anomaly scores are close to 0 after normalizing them across all sequences. ECVC and edit distance detection techniques are not affected by this problem as their anomaly scores are normalized per sequence rather than across all sequences; accordingly, both techniques yield better results than detection based on 2-grams. The results obtained for the Hadoop data set provide more insights into the problems arising from a dominating anomaly class that we mentioned in the previous section. Specifically, the highest F1 scores are achieved when the threshold is 0 and all instances are detected as anomalous, even though TNR is 0\% at that point. While TNR increases for higher thresholds, the F1 score decreases as TPR drops. All detectors yield comparatively low precision on the ADFA data set across all thresholds; 2-gram detection has high precision for high thresholds, but this is due to the fact that only few true positives are found. Overall, the diversity of the plots suggests that parameter tuning needs to be carried out on each data set separately to obtain optimal results.

\subsubsection{Event-based detection} \label{eventbased}

\begin{figure}
	\centering
	\includegraphics[width=1\columnwidth]{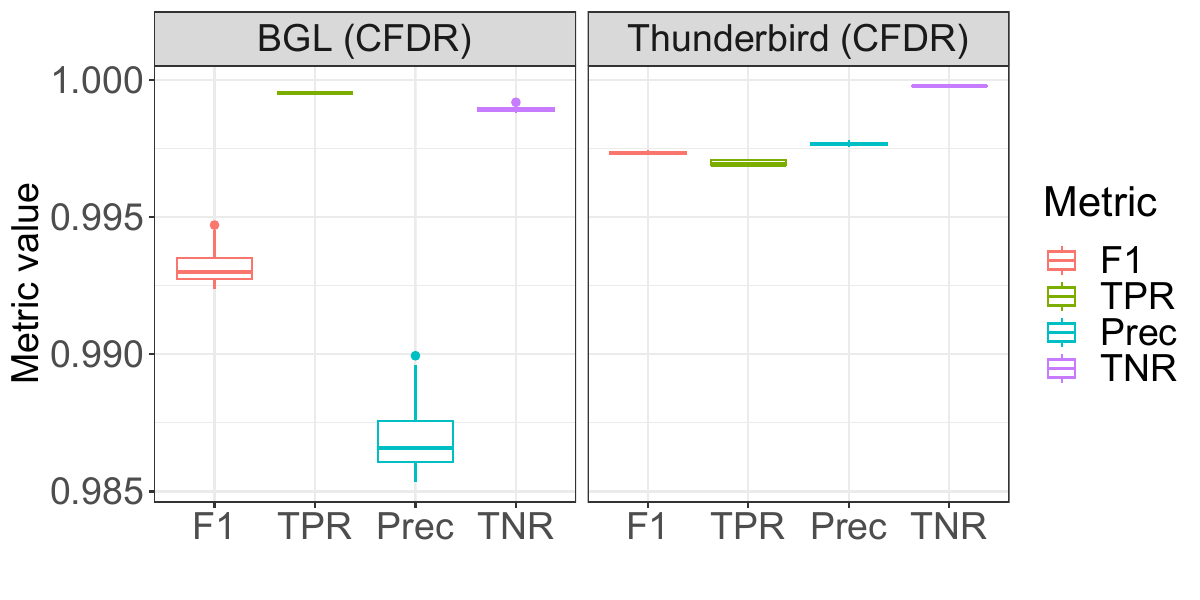}
	\caption{Evaluation results for event-based detection on BGL and Thunderbird data sets.}
	\label{fig:fone_events}
\end{figure}

Since single events are labeled in the BGL and Thunderbird data sets, we also apply the detection for new events on these data sets analogous to sequences. As we already discussed in Sect. \ref{bgl} and Sect. \ref{thunderbird}, almost all anomalous events belong to different types than normal events; thus, it is generally easy to detect anomalies, but false positives may be problematic in case that the training data set is not large enough to cover all normal events, which are subsequently reported in the detection phase. However, Fig. \ref{fig:fone_events} shows that very high detection scores and TNR of around 99.9\% is achieved in both data sets, indicating that 1\% is a sufficiently large training sample for this type of detection.

\section{Discussion} \label{discussion}

This section contains the discussion of our analysis and evaluation results. We first focus on the data sets themselves before moving to the way the data is used in evaluations. We end this section with some recommendations for future work.

\subsection{Appropriateness of Data Sets}

\begin{table*}[]
	\footnotesize
	\caption{Anomaly manifestations and identified issues in commonly used data sets}
	\label{tab:datasets}
	\begin{tabular}{llll}
		\textbf{Data set} & \textbf{Anomaly manifestations} & \textbf{Version} & \textbf{Issues} \\ \hline
		\multirow{6}{*}{HDFS} & \multirow{6}{*}{\makecell[l]{ - New event types \\ - Sequence lengths \\ - Event counts}} & Xu et al. \cite{xu2008mining} & \makecell[l]{ - The majority of anomalies is straightforward to detect. \\ - Some sub-sequences are randomly ordered, increasing sequence complexity.} \\ \cline{3-4} 
		&  & LogDeep & \makecell[l]{ - Same as for the original data set published by Xu et al. \\ - Training data are not a representative sample and yield excessively high detection rates. \\ - Timestamps for events cannot be retrieved as sequence identifiers are missing.} \\ \cline{3-4} 
		&  & Loghub \cite{he2020loghub} & \makecell[l]{ - Same as for the original data set published by Xu et al. \\ - Some lines are missing for unknown reasons.} \\ \hline
		\multirow{3}{*}{BGL}  & \multirow{3}{*}{\makecell[l]{ - New event types}} & CFDR \cite{oliner2007supercomputers} & \makecell[l]{ - Anomalies only affect certain event types and can be detected without sequences. \\ - Sometimes analyzed with sliding windows despite availability of sequence identifiers.} \\ \cline{3-4} 
		& & Loghub \cite{he2020loghub} & \makecell[l]{ - Same as for the original data set provided by CFDR. \\ - Some lines are missing or have been relabeled for unknown reasons.} \\ \hline
		Thunderbird  & \makecell[l]{ - New event types} & CFDR \cite{oliner2007supercomputers} & \makecell[l]{ - Anomalies only affect certain event types and can be detected without sequences.} \\ \hline
		OpenStack  & \makecell[l]{ - None} & Loghub \cite{he2020loghub} & \makecell[l]{ - Anomalies do not manifest as changes of event sequences or event inter-arrival times.} \\ \hline
		Hadoop  & \makecell[l]{ - New event types (few) \\ - None (most)} & Loghub \cite{he2020loghub} & \makecell[l]{ - The majority of anomalies do not manifest as changes of event sequences.} \\ \hline
	\end{tabular}
\end{table*}

To be suitable for anomaly detection evaluations, data sets need to meet characteristics that fit the type of detection. Given that the data sets analyzed in this study are widely used for the evaluation of deep learning anomaly detection techniques that ingest the data as sequences of event types and possibly event parameters, one would expect that changes in sequential patterns is exactly how the anomalies manifest in these data sets. This claim is supported by the fact that authors synthetically inject noise as part of evaluations by randomly shuffling sub-sequences as well as addition and deletion of certain events \cite{zhang2019robust}. However, the results of our study suggest that the link between anomalies and sequential patterns is less pronounced than expected. 

These findings are also reflected in Table \ref{tab:datasets}, which contains the answers to our research questions. In particular, we state the main types of anomaly manifestations for each of the reviewed data sets and describe identified issues and drawbacks for versions of publicly available data sets. As visible in the table, anomalies generally do not change sequential patterns and a major part of them is straightforward to differentiate from normal instances with simple detection methods.

The HDFS data set, which is the most popular data set in this research area \cite{landauer2023deep}, involves anomalies that manifest as new event types that do not occur within normal instances, unusually short sequence lengths, and event occurrence frequencies that do not require ordered sequences. In fact, random permutations of events that are generated simultaneously and not related to anomalies complicate sequence-based anomaly detection compared to other analysis techniques that are robust against permutations, such as count vector clustering. Another indicator that sequential patterns are less relevant in the HDFS data set than expected is that the ground truth of that data set was generated by clustering unordered count vectors \cite{xu2008mining}.

Our study further shows that a majority of the anomalies are straightforward to identify even for very simple detection approaches, which are capable of achieving competitive detection rates on the HDFS data set. While approaches such as Deeplog \cite{du2017deeplog} apply advanced detection models, a major part of their correctly detected instances thus implicitly relies on simple detection of new events \cite{le2022log} or detection of short sequences as a result of padding, i.e., augmenting short sequences with additional event types that make them more likely to be detected \cite{himler2023towards}. Unfortunately, this means that evaluation metrics reported on the HDFS data set give the misleading impression that a majority of anomalies are disclosed by the complex sequence-based detection techniques, even though they make up a much smaller share.

Anomalies in the BGL and Thunderbird data sets primarily manifest as events corresponding to types that never occur in normal data. Accordingly, for the simple task of identifying single events as anomalies, it is not required to analyze logs as sequences. In addition, grouping log data into sequences is often carried out using sliding windows \cite{chen2021experience}, which is problematic when different processes interleave \cite{landauer2023deep}. However, due to the length and complexity of these data sets, they appear well suited for many types of log analysis, such as automatic parser generation \cite{landauer2020system} or word embedding to compensate evolution of log messages \cite{zhang2019robust}. 

Both Hadoop and OpenStack data sets involve a high fraction of identical event sequences in normal and anomalous classes. No other artifacts suitable for anomaly detection in these data sets were identified in course of this study. We therefore advise against using these data sets for evaluation of anomaly detection techniques. 

Our study suggests that the ADFA data set is a promising alternative to aforementioned data sets, because anomalies are not detected by simple techniques such as new events or sequence lengths. In course of our evaluation study we were able to determine that some of the anomaly classes are easier to detect than others as suggested by related work \cite{xie2014evaluating}. Moreover, system call logs form ordered sequences that only involve discrete event types, which means that the composition of parsing templates has no influence on detection performance. However, we point out that our experiments are not suitable to confirm that the anomalies in the ADFA data set indeed manifest as changes of sequential patterns, which is a task we leave for future work.

Just because a data set is widely used in scientific publications does not necessarily mean that it is automatically a good choice. When Creech et al. \cite{creech2013generation} published the ADFA data set in 2013, they attempted to replace data sets such as KDD99 that were already criticized and considered outdated at that time \cite{siddique2019kdd}. Nonetheless, the data sets were still widely used in the research community many years later \cite{khraisat2019survey}. This shows that researchers are drawn towards data sets that are convenient to use (e.g., because they are labeled, sufficiently large, or available in pre-processed format) and accepted as a benchmark data set in the community, even though they are not ideal for evaluations. We thus expect that despite our findings, data sets such as the HDFS data set will continue to be used in the future unless superior alternatives are proposed.

\subsection{Appropriateness of Evaluations}

A major issue with most published evaluations is that the results are hardly reproducible or comparable. Since many authors do not publish their code, parameters of the algorithms, and data in a way that enables others to recreate the same results as stated in the respective paper, it is difficult to obtain an accurate overview of the detection capabilities achieved by state of the art approaches. Even though some authors re-implement well-known models \cite{chen2021experience, le2022log}, the evaluation results hardly ever align across multiple papers even though the same detection techniques are applied on the same data sets. 

One of the contributing factors for this issue is that authors usually fine-tune model parameters and repeat evaluation runs multiple times only to report the best results \cite{chen2021experience}. Unfortunately, this makes it difficult to comprehend the variance of the detection scores and the influence of selected thresholds. In addition, we observed that splits between training and test data vary strongly across different evaluations, including 1\% \cite{du2017deeplog}, 20\% \cite{catillo2022autolog}, 50\% \cite{hashemi2021onelog}, and 80\% \cite{le2022log, chen2021experience}. The choice of sampling strategies for the training data has also been shown to have a strong impact on the detection performance, in particular, the size of the training data and whether samples are drawn randomly from the whole data set (as we do in our study) or only the chronologically earliest samples are taken \cite{le2022log}. 

Another issue is that some papers only measure precision and recall to compute the F1 score \cite{guo2021logbert}, but omit false positive rate or true negative rate. Given that anomaly detection data sets are highly imbalanced, it is important to consider either of these metrics to avoid misinterpretation of results as we demonstrate on the Hadoop data set in Sect. \ref{seq}.

Anomaly detection techniques that leverage deep learning and neural networks generally have a lower explainability of classification results than conventional machine learning methods \cite{wolsing2022can}. Unfortunately, this also means that the actual reasons why instances are reported as anomalies can often go unnoticed. Accordingly, it is important to come up with evaluation methodologies and fitting data sets that are capable of demonstrating the advantages of these deep learning models in a clear way, in particular, to justify the higher runtime and computational effort in contrast to conventional machine learning methods \cite{chen2021experience}.

\subsection{Recommendations \& Future Work}

Based on the problems we identified in the previous sections, we formulate a set of recommendations to be addressed by future works.
\begin{enumerate}
	\item Create new data sets that specifically support evaluation of sequence-based detectors. System call logs such as the ADFA log data set appear beneficial as they are ordered, easy to group into sequences, avoid the need for parsing, and enable anomaly injection by executing adverse functions on the host where logs are collected.
	\item Repeat evaluation runs multiple times and report scores with variances. Presenting only the results from the best run with fine-tuned parameters causes that detection capabilities appear better than they are.
	\item Ensure that data sets are suitable for evaluation. In particular, the way anomalies manifest in the data should be verified and explained beforehand. 
	\item Use simple detection techniques fitting to the anomaly manifestations as baselines for comparison.
	\item Ensure reproducibility of reported results. This involves publishing all code and data that is necessary to repeat the conducted experiments and confirm the presented results. Moreover, relevant settings such as sampling strategies, splits between training and test data sets, and model parameters should be stated and their respective influence discussed in the paper.
\end{enumerate}
Our simple detection methods used in this paper focus on event occurrences, i.e., event timestamps are the only contextual information derived from the logs other than sequences of events. However, anomalies often manifest in event parameters, and incorporating them in the detection procedure is thus a reasonable approach. For example, simple parameter-based detection could leverage new value occurrences in categorical event parameters similar to our detection of new event types.

Finally, our study focuses on semi-supervised anomaly detection, where only normal data is used for training. However, supervised classification of anomalies is also an actively researched field that leverages the data sets used in this paper \cite{landauer2023deep}. We therefore plan to adopt our simple detection techniques for supervised anomaly detection and classification of anomalies into their respective classes. As this is considered out of scope for this paper, we leave this task for future work.

\section{Conclusion} \label{conclusion}

Quality and appropriateness of data sets are crucial for sound and representative evaluations of anomaly detection techniques. In this paper, we analyzed five log data sets that are commonly used in state of the art and one additional data set from security research for the purpose of determining whether they are suitable for the evaluation of sequence-based detection algorithms. While these algorithms are primarily designed to recognize changes of sequential patterns, such as log event types generated in different order than during normal operation, our analysis suggests that these artifacts hardly occur as part of anomalies. In the HDFS data set, for example, shuffled sub-sequences in event executions result from simultaneously generated events rather than anomalies. Moreover, anomalous sequences sometimes do not differ from normal behavior at all, rendering some data sets unusable. We tested the most suitable data sets with a small set of simple detection techniques that base on the detection of new events, sequence lengths, count vector similarity, sub-sequence similarity, and event timing. Our evaluation results suggest that these simple detectors are able to achieve competitive detection rates compared to advanced approaches from state of the art, further indicating that high detection rates are easy to achieve in some data sets. To counteract these issues, we recommend to work on new data sets that are specifically designed to include sequential anomalies and improve evaluation methodologies to avoid that misleading results are obtained for proposed detection approaches.

\section*{Acknowledgments}

This work was partly funded by the European Defence Fund (EDF) project AInception (101103385) and the FFG project PRESENT (FO999899544).



\bibliographystyle{IEEEtran}
\bibliography{bibliography}

 
%


\begin{IEEEbiography}[{\includegraphics[width=1in,height=1.25in,clip,keepaspectratio]{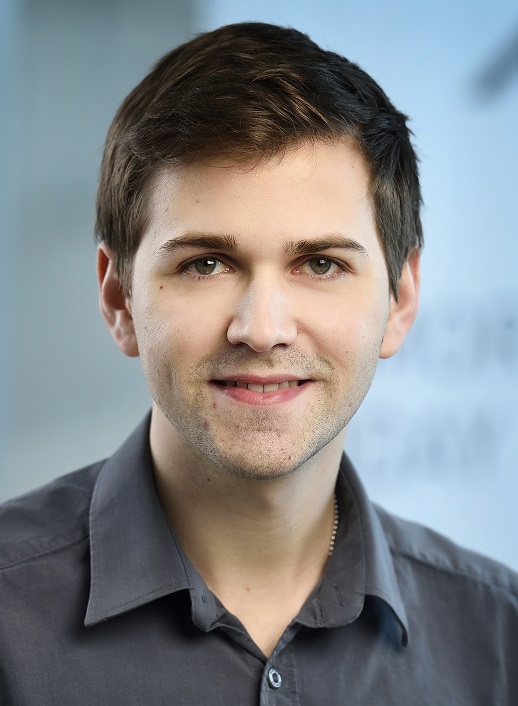}}]{Max Landauer}
	joined the Austrian Institute of Technology in 2017 and is currently employed as a scientist in the Cyber Security research group. His main research interests are anomaly detection, cyber threat intelligence, log data analysis, and cyber security testbeds. Max obtained his master’s degree in Computer Science in 2018 and finished his Ph.D. studies in 2022 at the Vienna University of Technology.
\end{IEEEbiography}

\begin{IEEEbiography}[{\includegraphics[width=1in,height=1.25in,clip,keepaspectratio]{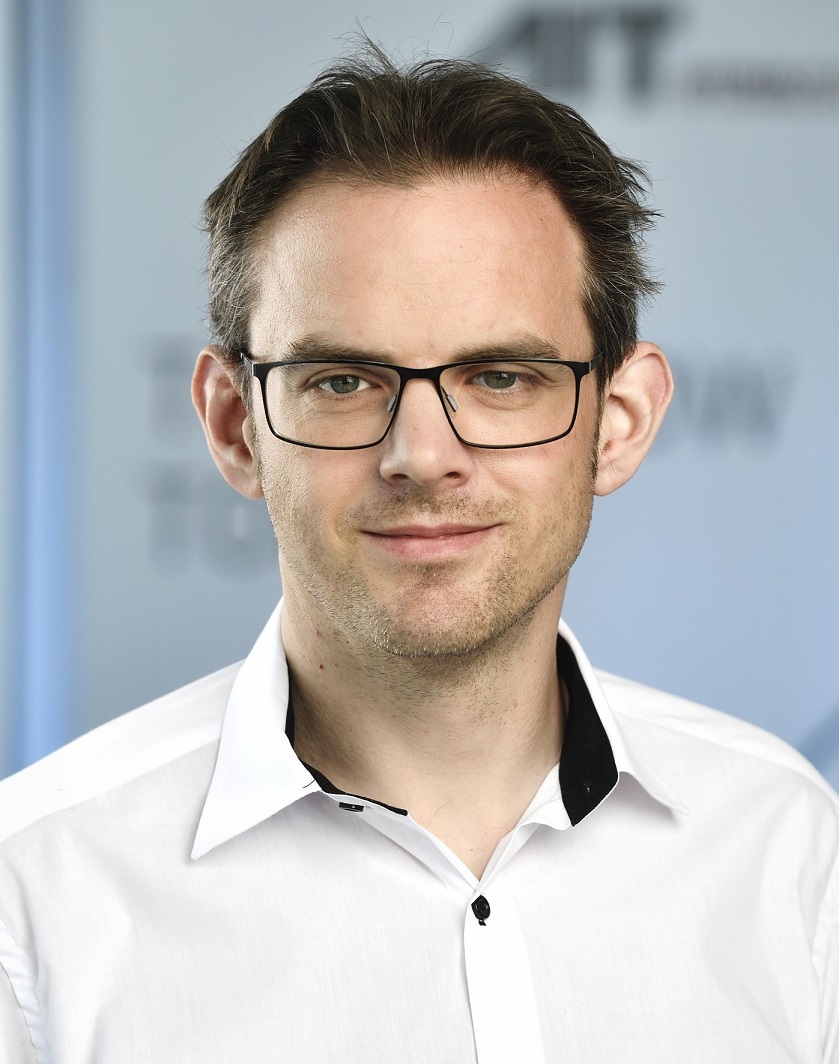}}]{Florian Skopik}
	is head of the cybersecurity research program at the Austrian Institute of Technology. His main interests are centered on critical infrastructure protection and intrusion detection. Florian received a Ph.D. in computer science from the Vienna University of Technology. He is a member of various conference program committees (e.g., ACM SAC, ARES, CRITIS), editorial boards, and standardization groups, such as ETSI TC Cyber, IFIP TC11, and OASIS CTI. He is a Senior Member of IEEE.
\end{IEEEbiography}

\begin{IEEEbiography}[{\includegraphics[width=1in,height=1.25in,clip,keepaspectratio]{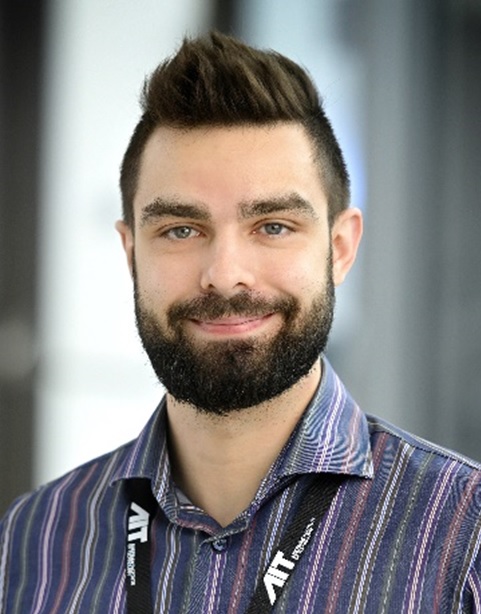}}]{Markus Wurzenberger}
	is a scientist and project manager at the Austrian Institute of Technology. Since 2014 he is part of the Cyber Security Research Group of AIT’s Center for Digital Safety and Security. His main research interests are log data analysis with focus on anomaly detection and cyber threat intelligence. Markus obtained a Ph.D. in computer science in 2021. In 2015 Markus obtained his master’s degree in Technical Mathematics at the Vienna University of Technology.
\end{IEEEbiography}

\vfill

\end{document}